\documentclass[pdflatex,sn-mathphys-num]{sn-jnl}% Math and Physical Sciences Numbered Reference Style 
\usepackage{graphicx}
\usepackage{graphicx}%
\usepackage{multirow}%
\usepackage{amsmath,amssymb,amsfonts}%
\usepackage{amsthm}%270033
 \usepackage{array}
\usepackage{mathrsfs}%
\usepackage[title]{appendix}%
\usepackage{xcolor}%
\usepackage{textcomp}%
\usepackage{manyfoot}%
\usepackage{booktabs}%
\usepackage{algorithm}%
\usepackage{algorithmicx}%
\usepackage{algpseudocode}%
\usepackage{listings}%
\usepackage{subfigure}
\usepackage{float}
\usepackage{lineno}
\usepackage{color}
\usepackage{url}
\usepackage{lipsum}
\usepackage{hyperref}%
\usepackage{subcaption}

\usepackage[defaultcolor=red]{changes}%
\theoremstyle{thmstyleone}%
\theoremstyle{thmstyletwo}%

\theoremstyle{thmstylethree}%

\begin{document}

\title[Article Title]{\textbf{Reconstructing Multi-Scale Physical Fields from Extremely Sparse Measurements with an Autoencoder–Diffusion Cascade}}


\author[1]{\fnm{Letian} \sur{Yi}}

\author[2]{\fnm{Tingpeng} \sur{Zhang}}

\author[1]{\fnm{Mingyuan} \sur{Zhou}}

\author[3]{\fnm{Guannan} \sur{Wang}}

\author[2,3,4]{\fnm{Quanke} \sur{Su}}

\author*[1,2,4]{\fnm{Zhilu} \sur{Lai}}\email{zhilulai@ust.hk}

\affil[1]{\orgdiv{Internet of Things Thrust}, \orgname{The Hong Kong University of Science and Technology (Guangzhou)},  \city{Guangzhou},  \country{China}}

\affil[2]{\orgdiv{Intelligent Transportation Thrust}, \orgname{The Hong Kong University of Science and Technology (Guangzhou)},  \city{Guangzhou},  \country{China}}

\affil[3]{\orgdiv{Marine Hydrodynamic Research Facility}, \orgname{The Hong Kong University of Science and Technology (Guangzhou)},  \city{Guangzhou},  \country{China}}

\affil[4]{\orgdiv{Department of Civil and Environmental Engineering}, \orgname{The Hong Kong University of Science and
Technology}, \state{Hong Kong SAR}, \country{China}}


\abstract{Extreme sensor sparsity makes full-field reconstruction a fundamentally ill-posed problem in scientific sensing, where the goal is to infer physical fields from sparse measurements. In this regime, the posterior is severely underconstrained and inherently multimodal, making its approximation highly ill-conditioned. Specifically, deterministic mappings collapse uncertainty, direct conditional learning cannot cover the space of possible observation-conditioned solutions, and likelihood-guided sampling becomes highly sensitive to noise and sensor configurations. These limitations result in unstable posterior estimates and highlight the need for modeling uncertainty in a structural manner. To this end, we propose Cascaded Sensing (Cas-Sensing), a hierarchical framework that restructures posterior inference across scales. Rather than modeling the full-field posterior directly, Cas-Sensing first resolves global structural ambiguity through a deterministic coarse-stage estimator. A neural-operator-based functional autoencoder, trained with masked inputs, maps sparse observations to a coarse-scale structural field, acting analogously to a maximum a posteriori estimator that selects the dominant global configuration. This structural anchor fixes the principal degrees of freedom of the posterior and transforms the problem into a better-conditioned residual inference task. A conditional diffusion model then learns only the refined-scale residual distribution, confining sampling to a stable neighborhood of plausible solutions and suppressing competition among observation-consistent modes. To enhance robustness under varying sensing conditions, we introduce mask-cascade training, which exposes the model to diverse sparse observation patterns through intermediate coarse reconstructions. During inference, manifold-constrained guidance enforces observation consistency as a refinement mechanism rather than a global mode-selection process. Through this staged design, Cas-Sensing combines deterministic structural inference with stochastic refinement, yielding a structured approximation of the posterior that remains stable under sparse and noisy observations. Experiments on simulated and real-world datasets demonstrate consistent generalization across sensor layouts, sparsity levels, noise conditions, and unseen geometric boundary configurations, producing stable and physically consistent reconstructions in both regular-domain fields and physical systems with internal boundaries. These results establish Cas-Sensing as a principled framework for scientific sensing, highlighting the necessity of structured uncertainty modeling in ill-conditioned inverse problems.}




\maketitle
\section{Introduction}\label{sec:introduction}

Although recent advances in sensor technology have improved the ability to monitor complex physical systems, dense observations remain prohibitively expensive in many applications due to the high cost of sensor deployment and maintenance. Therefore, spatial field reconstruction from limited local sensor information is a major challenge of complex physical systems, such as astrophysics \cite{akiyama2019first}, geophysics \cite{kondrashov2006spatio,carrassi2018data,manohar2018data}, atmospheric science \cite{alonso2010novel,mishra2014compressed}, and fluid dynamics \cite{vinuesa2023transformative,buzzicotti2023data}. 
In the regime of extremely sparse and random sensing, recovering multi-scale full-field states is not merely a missing-data interpolation task, but a severely underdetermined inverse problem in which the same observations may correspond to multiple physically plausible full-field solutions \cite{hadamard2014lectures,breiding2023algebraic}. The central objective is not to recover a single deterministic field, but to characterize the posterior distribution of feasible full-field states conditioned on sparse observations. In scientific sensing, this posterior can vary substantially with observation patterns and noise statistics, and under extreme sparsity even small changes in sensing configurations or noise characteristics may lead to large changes in the inferred solutions. Robust posterior modeling under extreme sparsity thus becomes a central challenge for scientific sensing.

Existing methods for sparse field reconstruction can be broadly understood according to how they handle this posterior ambiguity. One class of methods suppresses or collapses the ambiguity into a single deterministic solution. Compressed sensing belongs to this category: it reconstructs full-field data from undersampled observations by assuming that the target signal is sparse in a prescribed transform domain and solving an optimization problem under this strong prior \cite{candes2006robust}. To enhance reconstruction performance, compressed sensing variants combined with deep learning have been proposed, aiming to automatically learn effective sparse representations and reconstruction mappings in a data-driven manner \cite{sun2017intelligent,shi2019image,ni2020deep,xu2022sparse}. While these methods have achieved useful progress, their effectiveness still relies on restricting the solution space through strong priors or deterministic mappings, and they often struggle when the underlying physical fields are non-sparse, strongly multi-scale, or outside the assumed reconstruction setting.

A similar limitation appears in many recent deep learning methods that directly establish an end-to-end mapping between sparse measurements and the corresponding full-field data \cite{fukami2021global,fan2025vitae,li2025novel,ghazijahani2025spatial}. When encountering unseen sparse measurements—such as data acquired from a different sensor deployment configuration, or when facing new noise levels, these models often fail to produce accurate reconstructions and require retraining to adapt to the new scenarios. As a result, the applicability of these methods is generally limited to controlled or narrowly defined settings. They often struggle with generalization when applied to complex, real-world environments. From an inverse problem perspective, the limitation of end-to-end mappings is fundamentally rooted in the intrinsic uncertainty and non-uniqueness of sparse reconstruction, where the same sparse measurements may correspond to multiple physically plausible full-field solutions. This ambiguity is a defining characteristic of ill-posed inverse problems and cannot be resolved by deterministic regression alone \cite{bell1978solutions,arridge2019solving}. As a result, learning a single deterministic mapping inevitably collapses the underlying solution space, often leading to instability, overfitting to specific sensing configurations, and poor generalization when observations become sparser or deviate from the training distribution.

In principle, probabilistic generative modeling is a more appropriate paradigm because it aims to represent and sample from the conditional distribution of feasible solutions, rather than committing to a single point estimate. By learning data distributions rather than explicit input--output mappings, generative models, such as generative adversarial networks \cite{goodfellow2020generative} and diffusion models \cite{ho2020denoising,song2020score}, have shown remarkable capabilities in producing high-quality, diverse data, with broad applications in vision \cite{wang2023dr2,miao2024waveface,zhao2023towards}, biomedicine \cite{cao2024high,yu2020medical,nguyen2024sequence}, and scientific computing \cite{lienen2306zero,ruhling2023dyffusion,li2024synthetic}. These models can therefore generate unseen but statistically consistent samples.

\begin{figure}[H]
\centering 
\includegraphics[width=0.9\textwidth]{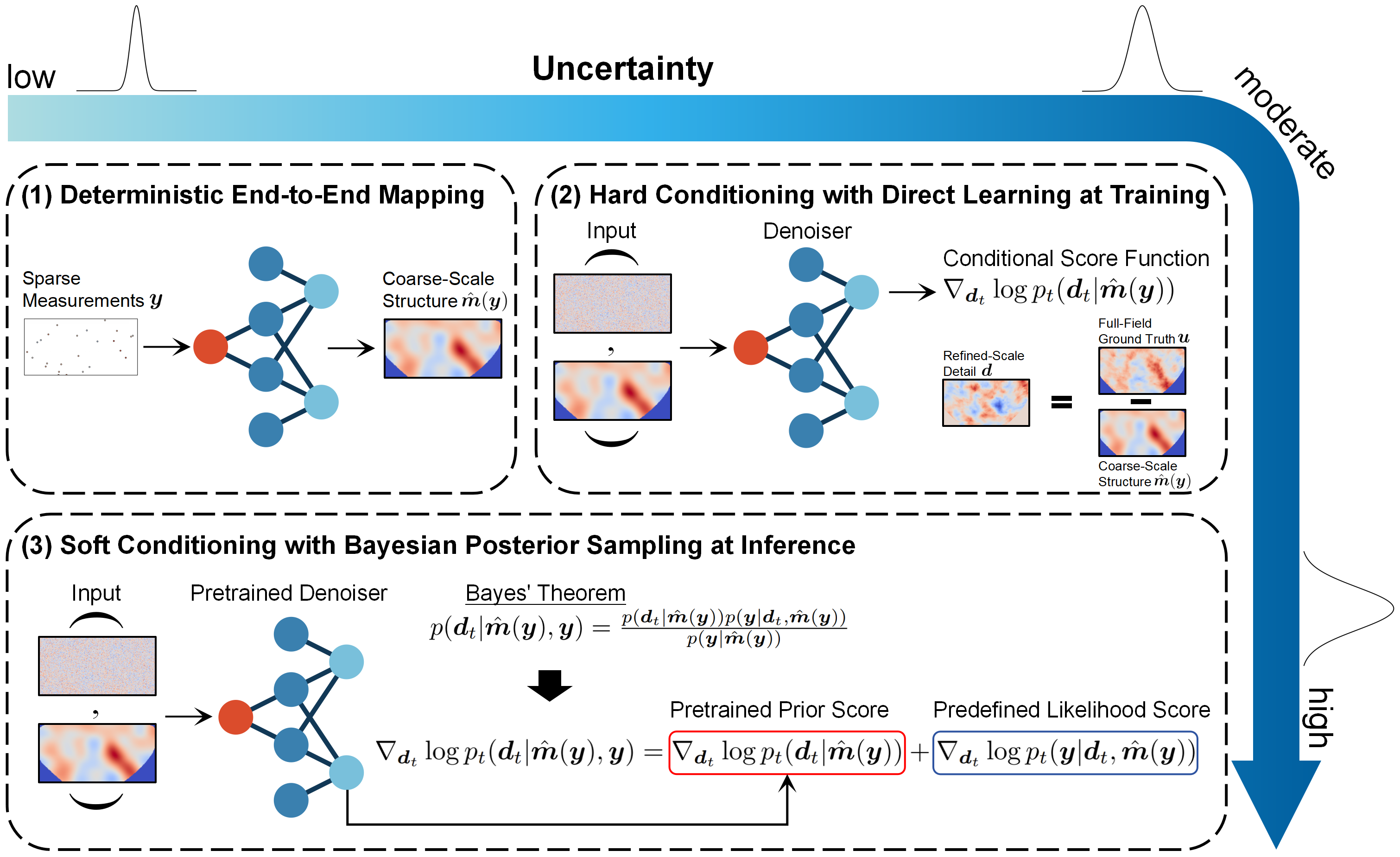}
\caption{Representative reconstruction strategies for sparse field recovery across increasing levels of uncertainty, and their correspondence to different subproblems in this work.}
\label{fig:technologies}
\end{figure}

Existing generative approaches for sparse field reconstruction can be broadly categorized into hard-conditioning and soft-conditioning paradigms, as illustrated in Fig.~\ref{fig:technologies}. These two paradigms can be viewed as different realizations of direct posterior modeling. In the hard-conditioning paradigm, the generative model is directly trained to represent a conditional distribution. This strategy is effective when the conditioning information is sufficiently restrictive and the uncertainty is relatively limited. One line of work learns supervised conditional distributions between paired low-resolution and high-resolution full-field data \cite{deng2019super,shu2023physics,shan2024pird}. Related efforts, such as state-observation augmented diffusion models, explicitly incorporate observations into training to learn posterior-related distributions \cite{li2025state}. Yet such approaches still rely on adequate training coverage of the state-observation relation. Under extremely sparse and randomly distributed measurements, the variability of sensing patterns is often too large for the training data to cover all sparse-to-full-field mappings, making direct posterior learning practically infeasible. Recent studies have combined physics-informed techniques \cite{raissi2019physics,karniadakis2021physics,yi2025transforming} with diffusion models for scientific inverse problems, using physical constraints to regularize the learned manifold and improve physical consistency \cite{shu2023physics,bastek2025physics}. However, in many practical sensing scenarios, the governing equations may be partially known, uncertain, computationally expensive to enforce, or unavailable at the target resolution. A related framework, FunDiff \cite{wang2025fundiff}, extends diffusion modeling to function spaces by combining a physics-informed functional autoencoder with latent diffusion. For inverse problems, FunDiff performs conditional diffusion based on encoded observations and uses PDE residual losses to regularize the functional representation. Although this latent-space formulation is effective for learning global functional structures across discretizations, its conditioning mechanism remains observation-encoding based. As a result, variability induced by extreme sparsity is still handled within a directly conditioned generation task.

In the soft-conditioning paradigm, a diffusion model is first pretrained on the original full-field data to provide a data-driven prior, while measurement information is incorporated at inference time through projection-based consistency steps \cite{choi2021ilvr,chung2022come,kadkhodaie2021stochastic} or Bayesian posterior sampling strategies \cite{du2024conditional,li2024learning,huang2024diffusionpde,chung2022diffusion,yao2025guided,shan2025red,dou2024diffusion}. This formulation avoids directly coupling the generative model with highly variable observation patterns during training, making it attractive for scientific sensing problems with substantial uncertainty. However, under extremely sparse observations, the inverse problem becomes severely ill-posed, and inference-time measurement constraints are often too weak to determine a reliable full-field solution from the unconditional prior. As a result, posterior sampling can fluctuate among multiple observation-consistent solutions supported by the learned prior. Our experiments show that this behavior does not necessarily reflect a stable multimodal posterior, but rather an ill-conditioned approximation in which weak observation constraints interact with the prior to produce unreliable mode selection. The problem becomes more pronounced when the sparse observations are noisy, as small measurement perturbations can induce substantial changes in the inferred posterior. Consequently, likelihood-guided sampling may generate fields that match the sparse observations but remain inaccurate over the full domain. Moreover, post hoc measurement enforcement can perturb the reverse trajectory away from the learned data manifold, introducing additional reconstruction errors \cite{chung2022improving}. Although manifold-constrained gradients \cite{chung2022improving} alleviate this issue, soft observation guidance remains a weak corrective mechanism rather than a structural constraint, and therefore cannot reliably stabilize posterior inference under extreme sparsity.

The discussion above indicates that generative reconstruction under extreme sparsity is limited by two coupled difficulties. First, sensing conditions can vary substantially across observation patterns, sensor layouts, and noise statistics, producing a diverse family of observation-conditioned reconstruction tasks and making generalization to unseen sensing configurations difficult. Second, extreme sparsity provides only weak structural constraints on the solution, leaving the posterior intrinsically ambiguous. As shown in our experiments, multiple solutions may be simultaneously consistent with the sparse observations and supported by the learned prior, leading to unstable sampling dynamics and unreliable mode selection. These difficulties reveal complementary limitations of existing paradigms. Fully hard-conditioning approaches require the model to learn a broad family of observation-conditioned mappings, which can hinder generalization under diverse sensing conditions. Purely soft-conditioning approaches avoid this training burden, but rely on weak inference-time guidance that is often insufficient to resolve competing solutions under extreme sparsity. Thus, the central issue is not simply whether to represent uncertainty, but how to structure it. Instead of learning the full posterior in a single step, reconstruction should be decomposed into simpler and better-conditioned subproblems, where components with different levels of variability are handled with appropriate conditioning strengths. In particular, resolving global structural ambiguity first and confining stochasticity to residual components provides a natural route to stabilizing posterior inference and reducing distributional complexity. A cascaded generative pipeline offers a principled strategy for this decomposition. By introducing intermediate representations, different stages of reconstruction can adopt different forms of conditioning while progressively organizing uncertainty. Related ideas have proven effective when direct one-shot generation is difficult \cite{nichol2021improved,saharia2022image,ho2022cascaded,oommen2024integrating}. In computer vision, cascaded generation is commonly implemented as multi-stage synthesis across increasing resolutions, where a base model captures coarse structure and subsequent models refine fine-scale details \cite{ho2022cascaded}. Compared with a single high-resolution generator, this design separates structural reconstruction from detail refinement and allows each stage to be tailored to its specific role.

\begin{figure}[H]
\centering 
\includegraphics[width=0.9\textwidth]{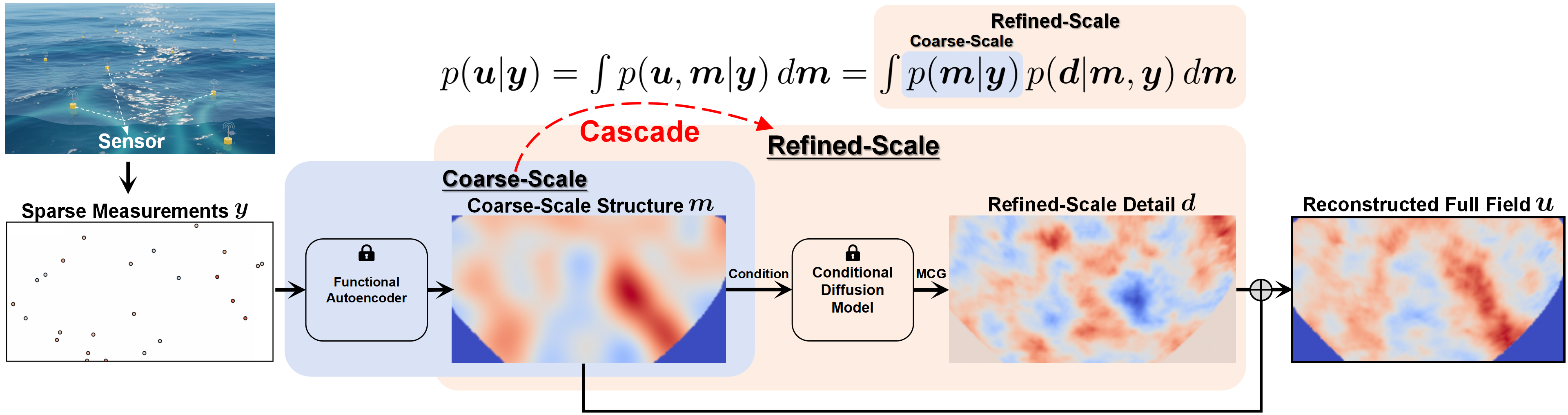}
\caption{The overview of Cascaded Sensing, where $\boldsymbol{d}=\boldsymbol{u}-\boldsymbol{m}$ and MCG denotes Manifold Constrained Gradient for imposing the measurement consistency.}
\label{fig:overview of Cas-Sensing}
\end{figure}

As shown in Fig.~\ref{fig:overview of Cas-Sensing}, we propose Cascaded Sensing (Cas-Sensing), an autoencoder--diffusion cascade for robust multi-scale field reconstruction from extremely sparse measurements. The key idea is to restructure posterior inference into two better-conditioned stages, so that variability in observation patterns and ambiguity among observation-consistent solutions are handled separately. First, a neural-operator-based functional autoencoder is trained with masked inputs \cite{bunker2024autoencoders} to deterministically map arbitrary sparse measurements to physically plausible coarse-scale fields. This stage converts diverse observation-conditioned reconstruction tasks into a unified coarse-structure estimation problem, thereby anchoring the dominant global degrees of freedom. The original full-field posterior is then transformed into a refined-scale detail distribution that is simpler and more concentrated. A conditional diffusion model is subsequently trained to model this detail distribution conditioned on the inferred coarse structure, rather than directly learning the full observation-conditioned posterior. This confines stochastic sampling to a neighborhood of the estimated global structure and reduces competition among multiple observation-consistent modes. To improve robustness to changing sensing configurations, we introduce Mask-Cascade Training (MCT), where random sparse masks are passed through the pretrained functional autoencoder to generate diverse coarse conditions and corresponding refined-scale details from the same ground-truth field. This exposes the diffusion model to condition variability induced by different observation patterns without requiring it to learn all observation-conditioned mappings explicitly. During inference, sparse measurements are incorporated through Manifold Constrained Gradient (MCG) guidance \cite{chung2022improving}. Since the global structural anchor has already been inferred, MCG acts mainly as a local refinement mechanism rather than a full-field mode selector, avoiding the unstable posterior dynamics of purely soft-conditioning approaches. Through this staged design, Cas-Sensing combines hard structural conditioning with soft observation-guided refinement, organizing uncertainty across scales and improving robustness to sparsity, noise, and varying sensing patterns. We further show that the same framework can be extended to physical fields with internal geometric boundaries, where the functional autoencoder provides transferable coarse structural conditions for boundary configurations not observed during training.

The remainder of the paper is organized as follows. In Section 2, we present the problem formulation and describe the proposed methodology in detail, including the functional autoencoder, the conditional diffusion model, and the proposed mask-cascade training approach. Section 3 demonstrates the effectiveness of Cas-Sensing through a series of experiments with simulated data of circular cylinder flow and Navier--Stokes vorticity flow, real-world data of ocean wave height and global ocean temperature. In Section 4, we discuss the advantages and limitations of Cas-Sensing. Finally, Section 5 concludes the paper and outlines directions for future work.

\section{Methods}\label{sec:methods}

\subsection{Problem formulation}\label{sec:problem formulation}

The problem of recovering the multi-scale full field $\boldsymbol{u}\in \mathbb{R}^m$ from a vector of sparse sensor measurements $\boldsymbol{y}\in \mathbb{R}^n$ can be formulated as modeling the conditional distribution $p(\boldsymbol{u}|\boldsymbol{y})$. However, under extreme sparsity, the posterior $p(\boldsymbol{u}|\boldsymbol{y})$ becomes highly sensitive to variations in observation patterns and noise statistics, making direct modeling difficult and brittle under changing sensing conditions. To address this challenge, we introduce an intermediate variable $\boldsymbol{m}$ into this probabilistic framework, thereby forming a cascaded pipeline.

We leverage the following hierarchical probabilistic decomposition via marginalization:
\begin{equation}
p(\boldsymbol{u}|\boldsymbol{y})
=\int p(\boldsymbol{u},\boldsymbol{m}|\boldsymbol{y}) \,d\boldsymbol{m}
= \int p(\boldsymbol{u}|\boldsymbol{m},\boldsymbol{y}) \, p(\boldsymbol{m}|\boldsymbol{y}) \, d\boldsymbol{m}
=\mathbb{E}_{\boldsymbol{m}\sim p(\boldsymbol{m}|\boldsymbol{y})}[\,p(\boldsymbol{u}|\boldsymbol{m},\boldsymbol{y})\,],
\end{equation}
where $\boldsymbol{m}$ represents the coarse-scale structure (the coarse-scale component) of $\boldsymbol{u}$. This formulation naturally decomposes the task into two subtasks: in principle, one may first model $p(\boldsymbol{m}|\boldsymbol{y})$ to infer the coarse-scale structure $\boldsymbol{m}$ from sparse observations $\boldsymbol{y}$, and then model $p(\boldsymbol{u}|\boldsymbol{m},\boldsymbol{y})$ to reconstruct the full field $\boldsymbol{u}$ by adding refined details to $\boldsymbol{m}$ conditioned on both $\boldsymbol{m}$ and $\boldsymbol{y}$. A strict realization of this decomposition would require learning a stochastic model for $p(\boldsymbol{m}|\boldsymbol{y})$, repeatedly sampling plausible $\boldsymbol{m}$ from it, and then training a conditional generative model for $p(\boldsymbol{u}|\boldsymbol{m},\boldsymbol{y})$ while marginalizing over $\boldsymbol{m}$, e.g., through Monte Carlo estimation.
At coarse scales, physical fields are dominated by low-frequency, smoothly varying components that can be represented with far fewer degrees of freedom than refined-scale fluctuations. As a result, reconstructing $\boldsymbol{m}$ from sparse observations -- i.e., modeling $p(\boldsymbol{m}|\boldsymbol{y})$ -- is generally better conditioned than directly reconstructing the full field.
Furthermore, compared to directly modeling $p(\boldsymbol{u}|\boldsymbol{y})$, modeling $p(\boldsymbol{u}|\boldsymbol{m},\boldsymbol{y})$ incorporates an additional condition, constraining the reconstruction of $\boldsymbol{u}$ to remain consistent with the coarse-scale structure $\boldsymbol{m}$. Moreover, the intermediate variable $\boldsymbol{m}$ serves not only as an additional structural condition, but also as a natural carrier through which uncertainty induced by sparse observations can be propagated hierarchically toward full-field reconstruction.

However, such a strict implementation is computationally expensive and practically cumbersome, since it requires explicitly learning and sampling from $p(\boldsymbol{m}|\boldsymbol{y})$ and propagating this stochasticity through a second-stage generative model. More fundamentally, under extreme sparsity, the posterior over coarse-scale structures itself can become highly ill-conditioned and sensitive to sparse observation perturbations. Strictly preserving the full stochasticity of $p(\boldsymbol{m}\mid\boldsymbol{y})$ may therefore inherit this instability, causing the subsequent generative process to fluctuate among competing structural modes. Moreover, directly learning $p(\boldsymbol{u}|\boldsymbol{m},\boldsymbol{y})$ would still require the second-stage model to cover the enormous variability induced by sparse observation patterns, which is impractical under extreme sparsity. Motivated by these limitations, Cas-Sensing introduces the intermediate representation $\boldsymbol{m}$ but adopts a stabilized approximation. Specifically, a functional autoencoder is trained to provide a deterministic estimate of coarse-scale structure from sparse observations, serving as a hard structural anchor that suppresses unstable coarse-scale posterior variability. The remaining variability is then partially reintroduced through the proposed Mask-Cascade Training (MCT) strategy, while sparse observations are incorporated at inference time through Manifold Constrained Gradient (MCG) guidance. In this way, Cas-Sensing preserves the hierarchical spirit of the probabilistic decomposition while avoiding both the prohibitive cost and the instability associated with explicitly modeling the full stochastic distribution $p(\boldsymbol{m}|\boldsymbol{y})$.

\subsection{Functional autoencoder for coarse-scale structure reconstruction}\label{sec:FAE}

\begin{figure}[H]
    \centering
    \includegraphics[width=0.9\linewidth]{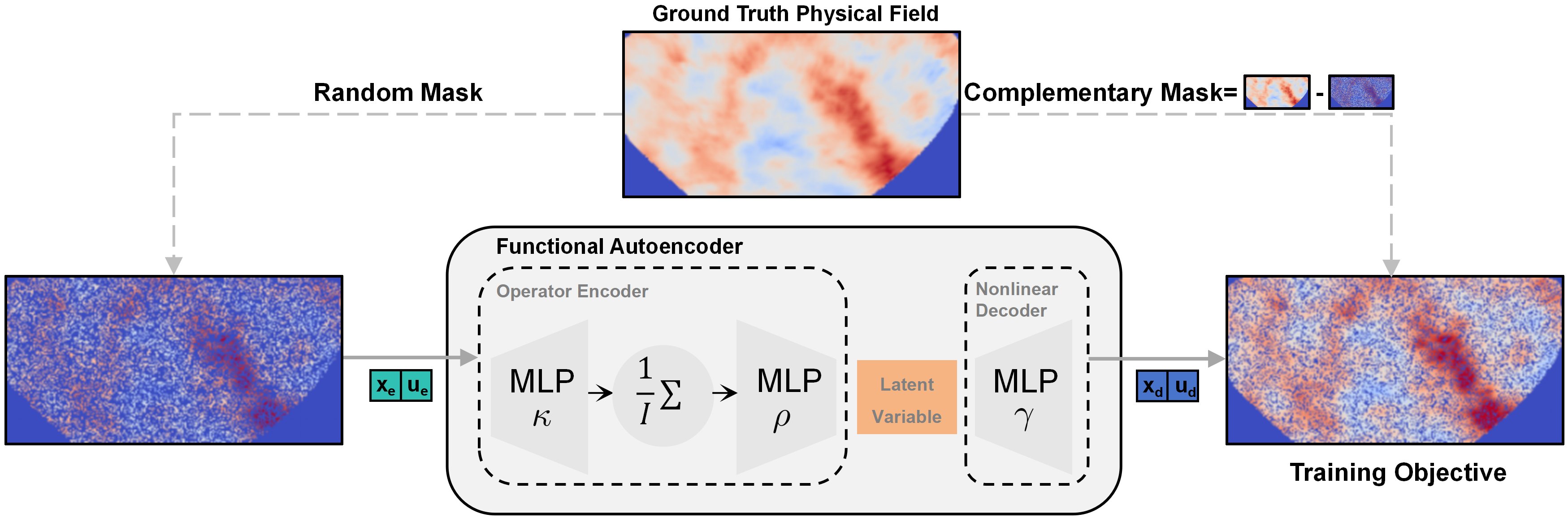}
    \caption{The architecture and masked training scheme of functional autoencoder. $x_e$ and $u_e$ denote the coordinates and field values fed into the encoder, while $x_d$ and $u_d$ represent the coordinates and reconstructed values produced by the decoder.}
    \label{fig:masked training of FAE}
\end{figure}

In practical engineering, sensors are often randomly placed (or some sensors are even movable), and many physical fields involve complex geometric boundaries, making field reconstruction from unstructured data challenging \cite{chai2020deep}. To address this, we adopt a neural operator-based functional autoencoder \cite{bunker2024autoencoders}, which enables coordinate-based input for neural networks \cite{regazzoni2024learning,krishnapriyan2023learning}. The functional autoencoder extracts latent features from unstructured data and robustly reconstructs a coarse-scale structural estimate from random and sparse inputs.

We model the inverse mapping from sparse observations $\boldsymbol{y}$ to the coarse-scale structure $\boldsymbol{m}$ using a deterministic autoencoder rather than a fully probabilistic coarse-level model (e.g., a variational autoencoder). The goal is not to claim that the true posterior $p(\boldsymbol{m}|\boldsymbol{y})$ is strictly deterministic, but to obtain a tractable point estimate of the coarse structural component that serves as a stable intermediate representation in the cascaded framework. Since coarse-scale patterns are smoother and have far fewer degrees of freedom than refined-scale fluctuations, such a point-estimate approximation is effective for providing a structural anchor. Accordingly, we write
\begin{equation}
    p(\boldsymbol{m}|\boldsymbol{y}) \approx \delta(\boldsymbol{m} - \hat{\boldsymbol{m}}(\boldsymbol{y}))
    \label{eq:dirac_delta}
\end{equation}
where $\delta(\cdot)$ is a Dirac delta function and $\hat{\boldsymbol{m}}(\boldsymbol{y})$ is the deterministic coarse-scale estimate produced by the autoencoder. This approximation is introduced for computational tractability, rather than to assert that all uncertainty in $p(\boldsymbol{m}|\boldsymbol{y})$ vanishes. The variability compressed by this point-estimate approximation is partially reintroduced later through mask-cascade training, while observation-dependent uncertainty is handled during sampling through observation-guided refinement.

\subsubsection{The architecture of functional autoencoder}\label{sec: OAE architecture}
In this study, we employ a neural-operator-based autoencoder built upon the framework introduced in \cite{bunker2024autoencoders} to reconstruct the coarse structural component from sparse observations. The encoder $\mathcal{E}$ maps a function $u:\Omega \rightarrow \mathbb{R}^m$ to a latent vector $z \in \mathbb{R}^{d_z}$. In developing the encoder–decoder architectures, we assume that $\mathcal{U}$ is a Banach space of functions that can be evaluated pointwise almost everywhere, with domain $\Omega \subset \mathbb{R}^d$ and codomain $\mathbb{R}^m$. In practice, we only have access to discretized representations $u(x) \in \mathcal{U}$, obtained by evaluating $u$ at a finite set of mesh points. Following the formulation in \cite{li2020neural,li2020fourier}, we parametrize the encoder as a kernel integral operator:
\begin{equation}\label{sec: pooling encoder}
    \mathcal{E}(u;\zeta)=\rho\left(\int_\Omega\kappa\left(x,u(x);\zeta_1\right)\mathrm{d}x;\zeta_2\right)\in\mathcal{Z}=\mathbb{R}^{d_z},
\end{equation}
with $\kappa: \Omega\times\mathbb{R}^m\times\zeta_1\rightarrow\mathbb{R}^l$ parameterized as a neural network with hidden layers of width $l=64$, using GELU activation, and with $\rho:\mathbb{R}^l\times\zeta_2\rightarrow\mathbb{R}^{d_z}$ parameterized as the linear layer $\rho(v;\zeta_2)=W^{\zeta_2}v+b^{\zeta_2}$, where $d_z$ denotes the dimension of latent vector $z$ and $\zeta=[\zeta_1,\zeta_2]$ denotes the parameters of encoder. We augment $x\in\Omega$ with 16 random Fourier features to enrich the coordinate representation and alleviate spectral bias in learning the encoder mapping \cite{tancik2020fourier,li2020fourier}. After discretizing the data as \(u=\{(x_i,u(x_i))\}_{i=1}^{I}\), we approximate the integral over \(\Omega\) by a normalized sum:
\begin{equation}
\{(x_i,u(x_i))\mid i=1,2,\ldots,I\}
\rightarrow
\rho\!\left(
\mathrm{pool}\!\left(
\{\kappa(x_i,u(x_i);\zeta_1)\mid i=1,2,\ldots,I\}
\right);
\zeta_2
\right).
\end{equation}
where pool denotes a pooling operation that is invariant to the order of its inputs—specifically, average pooling in our implementation. Owing to its permutation-invariant property, average pooling enables the functional autoencoder to process arbitrarily ordered and distributed sparse inputs. Moreover, it can adaptively attenuate high-frequency variations while preserving low-frequency components, which is consistent with the well-known smoothing effect of averaging operations. Since average pooling reduces the discrepancy between latent representations obtained from sparse and dense observations, the learned latent space becomes largely insensitive to the input sampling ratio, endowing the functional autoencoder with robust and accurate reconstruction capability under extreme sparsity.

The decoder $\mathcal{D}$ maps a latent vector $z \in \mathcal{Z}$ back to a function $\mathcal{D}(z;\psi): \Omega \rightarrow \mathbb{R}^m$. To realize this mapping, we parameterize $\mathcal{D}$ with a coordinate-based neural network $\gamma: \mathcal{Z} \times \Omega \times \Psi \rightarrow \mathbb{R}^m$, whose hidden layers employ GELU activations. Accordingly, we have
\begin{equation}\label{sec: nonlinear decoder}
\mathcal{D}(z;\psi)(x) = \gamma(z, x; \psi) \in \mathbb{R}^m.
\end{equation}
In this construction, each spatial coordinate $x \in \Omega$ is enriched with 16 random Fourier features. Inspired by DeepONet \cite{lu2021learning}, the proposed architecture enables the decoded function $\mathcal{D}(z;\psi)$ to be evaluated on arbitrary meshes, with the computational cost scaling linearly in the number of discretization points.

\subsubsection{Training objective and masked training}\label{sec: OAE loss and masked training}
The central aim of training a functional autoencoder is to compress functional data $u$ into a low-dimensional latent code while retaining sufficient information to accurately reconstruct the original signal. Formally, the optimization problem is expressed as
\begin{equation}\label{eq:FAE loss} 
\mathcal{L}(\zeta,\psi)=\frac{1}{2}||\mathcal{D}(\mathcal{E}(u;\zeta);\psi)-u||^2_2+\beta||\mathcal{E}(u;\zeta)||^2_2, \end{equation}
where the second term, $||\mathcal{E}(u;\zeta)||_2^2$ with coefficient $\beta$, plays the role of regularization, ensuring that the latent representation is smooth and structured.

A widely adopted strategy to further enhance autoencoder performance is self-supervised masked training, in which the model learns to recover missing portions of data from partially observed inputs \cite{he2022masked}. This training paradigm not only accelerates convergence but also improves generalization. More importantly, it equips the network with the ability to reason about incomplete data, which is particularly advantageous when dealing with sparse or irregular samples \cite{devlin2019bert,liu2021robustly,raffel2020exploring}. In our framework, we employ the complement mask training approach proposed in \cite{bunker2024autoencoders}, which exploits the flexibility of discretising both encoder and decoder on arbitrary computational meshes. Specifically, consider a discretised function sample $\boldsymbol{u}=\{(x_i,u(x_i))\}_{i=1}^m$. At each training iteration, two index subsets $\mathcal{I}_\text{enc}$ and $\mathcal{I}_\text{dec}$ of $\mathcal{I}=\{1,\ldots,m\}$ are randomly drawn to form $\boldsymbol{u}_\text{e}=\{(x_i,u(x_i))|i\in\mathcal{I}_\text{enc}\}$ and $\boldsymbol{u}_\text{d}=\{(x_i,u(x_i))|i\in\mathcal{I}_\text{dec}\}$. In practice, $\mathcal{I}_\text{enc}$ is chosen as a random subset of $\mathcal{I}$, while $\mathcal{I}_\text{dec}$ is set to its complement, i.e., $\mathcal{I}_\text{dec}=\mathcal{I}\backslash \mathcal{I}_\text{enc}$. The ratio $r_\text{enc} = |\mathcal{I}_\text{enc}|/|\mathcal{I}|$ is treated as a tunable hyperparameter. As depicted in Fig. \ref{fig:masked training of FAE}, the encoder processes $\boldsymbol{u}_\text{e}$, while the decoder is evaluated on the complementary mesh $\{x_i\}_{i\in \mathcal{I}_\text{dec}}$; the resulting output is then compared against $\boldsymbol{u}_\text{d}$.

\subsection{Conditional diffusion models for refined-scale detail generation}\label{sec: CDM}

In Cas-Sensing, the second stage is designed to generate refined-scale details conditioned on the recovered coarse-scale structures, while enforcing consistency with sparse measurements. To this end, we adopt a Denoising Diffusion Probabilistic Model (DDPM) as the generative backbone, as its Bayesian sampling interpretation allows observation-dependent constraints to be injected at inference time without retraining the model. 

After reconstructing the coarse-scale structures $\boldsymbol{m}$ with the functional autoencoder, the conditional diffusion model is trained to generate the refined-scale components defined as the detail $\boldsymbol{d} = \boldsymbol{u} - \boldsymbol{m}$. 
Modeling the detail significantly simplifies the learning task, since the refined-scale residual distribution is substantially less complex than the original full-field distribution and naturally constrains the reconstruction to remain close to the coarse-scale estimate. As discussed in Eq.~\eqref{eq:dirac_delta}, the deterministic coarse-scale estimate $\hat{\boldsymbol{m}}(\boldsymbol{y})$ induces the following tractable approximation to the cascaded probabilistic formulation:
\begin{equation}
\label{eq:cascade_eq}
p(\boldsymbol{u}\mid \boldsymbol{y})
= \int p(\boldsymbol{u}\mid \boldsymbol{m}, \boldsymbol{y}) \, p(\boldsymbol{m}\mid \boldsymbol{y}) \, d\boldsymbol{m}
\approx \int p(\boldsymbol{d}\mid \boldsymbol{m}, \boldsymbol{y}) \, \delta(\boldsymbol{m}-\hat{\boldsymbol{m}}(\boldsymbol{y})) \, d\boldsymbol{m}
= p(\boldsymbol{d}\mid \hat{\boldsymbol{m}}(\boldsymbol{y}), \boldsymbol{y}).
\end{equation}

Directly training the diffusion model to learn $p(\boldsymbol{d}\mid \hat{\boldsymbol{m}}(\boldsymbol{y}),\boldsymbol{y})$ is nevertheless undesirable: the variability compressed by the point-estimate approximation of $p(\boldsymbol{m}\mid \boldsymbol{y})$ would be inherited by the second-stage model, while the model would also need to cover the enormous variability of sparse sensing patterns. Therefore, rather than using Eq.~\eqref{eq:cascade_eq} as the direct Conditional Diffusion Model (CDM) training objective, we reinterpret it from a Bayesian perspective.

Specifically, for a given coarse-scale estimate $\boldsymbol{m}$, the detail posterior can be decomposed as
\begin{equation}
\label{eq:bayes_decomp}
p(\boldsymbol{d}\mid \hat{\boldsymbol{m}}(\boldsymbol{y}),\boldsymbol{y}) \propto p(\boldsymbol{y}\mid \boldsymbol{d},\hat{\boldsymbol{m}}(\boldsymbol{y}))\, p(\boldsymbol{d}\mid \hat{\boldsymbol{m}}(\boldsymbol{y})),
\end{equation}
where $p(\boldsymbol{d}\mid \hat{\boldsymbol{m}}(\boldsymbol{y}))$ is a detail prior conditioned on the coarse structure, and $p(\boldsymbol{y}\mid \boldsymbol{d},\hat{\boldsymbol{m}}(\boldsymbol{y}))$ is the observation-consistency likelihood. In our implementation, the CDM is used to learn $p(\boldsymbol{d}\mid\hat{\boldsymbol{m}}(\boldsymbol{y}))$ with hard conditioning, namely a detail prior conditioned only on plausible coarse structures, without explicitly accessing the corresponding sparse observation $\boldsymbol{y}$ during training. This decoupling prevents the generative model from being tied to specific observation patterns and improves generalization to unseen sensor configurations. Since the CDM is conditioned only on the coarse structural estimate \(\hat{\boldsymbol{m}}(\boldsymbol{y})\), it learns the residual distribution associated with the structure of \(\hat{\boldsymbol{m}}\), rather than the specific sparse observation pattern \(\boldsymbol{y}\) that produces it. Thus, the burden of generalization is shifted from enumerating sparse samplings to learning over a sufficiently diverse family of coarse conditions, which is precisely the role of the proposed Mask-Cascade Training (MCT) strategy.

At inference time, the observation-dependent likelihood term in Eq.~\eqref{eq:bayes_decomp} is incorporated through the Manifold Constrained Gradient (MCG), which injects measurement-consistency guidance into the reverse diffusion process. Therefore, the second stage of Cas-Sensing should be understood as learning a hard-conditioned detail prior with MCT-enhanced conditioning diversity, followed by MCG-based observation guidance that constrains the generated details to be consistent with the actual sparse measurements.

\subsubsection{Conditional denoising diffusion probabilistic model}\label{sec: DDPM}

Training of DDPMs consists of a diffusion process and a denoising process. Diffusion process transforms $\boldsymbol{d}_0$ from the real data distribution $\boldsymbol{d}_0 \sim p(\boldsymbol{d}_0)$ into a pure Gaussian noise $\boldsymbol{d}_t\sim \mathcal{N}(0,\boldsymbol{I})$ by successively applying the following Markov diffusion kernel:
\begin{equation}
    p(\boldsymbol{d}_t|\boldsymbol{d}_{t-1}) = \mathcal{N}\left( \boldsymbol{d}_t; \sqrt{1 - \beta_t} \, \boldsymbol{d}_{t-1}, \, \beta_t \mathbf{I} \right),
\end{equation}
where $\{\beta_t\}^T_{t=1}$ is a pre-defined or learned noise variance schedule. We adopt a linear noise schedule where the variance $\beta_t$ increases linearly from $1\times10^{-4}$ to $0.02$ over $T=1000$ diffusion steps. The marginal distribution at arbitrary timestep $t$ can be denoted as:
\begin{equation}
    p(\boldsymbol{d}_t|\boldsymbol{d}_0) = \mathcal{N}\left( \boldsymbol{d}_t; \sqrt{\bar{\alpha}_t} \, \boldsymbol{d}_0, \, (1 - \bar{\alpha}_t) \, \mathbf{I} \right),
\end{equation}
where $\bar{\alpha}_t = \prod_{s=1}^{t}\alpha_s, \alpha_t = (1 - \beta_t)$. When $T\rightarrow\infty,\ p(\boldsymbol{d}_t|\boldsymbol{d}_0)\approx\mathcal{N}(\mathbf{0}, \mathbf{I})$. For sampling, $\boldsymbol{d}_t = \sqrt{\bar{\alpha}_t}\boldsymbol{d}_0+(1 - \bar{\alpha}_t)\boldsymbol{\epsilon},\ \boldsymbol{\epsilon} \sim \mathcal{N}(\mathbf{0}, \mathbf{I})$.

Reversely, given $\boldsymbol{d}_t$, the denoising process aims to recover $\boldsymbol{d}_0$ by recursively learning the transition from $\boldsymbol{d}_{t-1}$ to $\boldsymbol{d}_t$, which is defined as the following Gaussian distribution:
\begin{equation}
q_\theta(\boldsymbol{d}_{t-1}|\boldsymbol{d}_t) = \mathcal{N}\left( \boldsymbol{d}_{t-1}; \boldsymbol{\mu}_\theta(\boldsymbol{d}_t, t), \boldsymbol{\Sigma}_\theta(\boldsymbol{d}_t, t) \right),
\end{equation}
where parameters $\theta$ are optimized by a denoising network $\boldsymbol{\epsilon}_\theta$ that predicts $\boldsymbol{\mu}_\theta(\boldsymbol{d}_t, t)=\frac{1}{\sqrt{\alpha_t}} \left( \mathbf{\boldsymbol{d}}_t - \frac{1 - \alpha_t}{\sqrt{1 - \bar{\alpha}_t}} \, \boldsymbol{\epsilon}_\theta(\mathbf{\boldsymbol{d}}_t, t) \right)$. As for $\boldsymbol{\Sigma}_\theta(\boldsymbol{d}_t, t)$, we use the closed-form posterior variance $\sigma_t^2 = \beta_t\frac{1 - \bar{\alpha}_{t-1}}{1 - \bar{\alpha}_t}$ as proposed in the original denoising diffusion probabilistic model formulation \cite{ho2020denoising}, instead of predicting it via neural network.

As discussed above, in the absence of a coarse-scale structural condition, posterior modeling under extreme sparsity becomes highly unstable, as weak measurement constraints cannot reliably resolve competing observation-consistent solutions. Therefore, the coarse-scale inference $\hat{\boldsymbol{m}}(\boldsymbol{y})$ is injected as the condition for the denoising process:
\begin{equation}
    q_\theta(\boldsymbol{d}_{t-1}|\boldsymbol{d}_{t},\hat{\boldsymbol{m}}(\boldsymbol{y}))=\mathcal{N}(\boldsymbol{d}_{t-1};\boldsymbol{\mu}_{\theta}(\boldsymbol{d}_{t},t,\hat{\boldsymbol{m}}(\boldsymbol{y})),\sigma^2_t\mathbf{I}),
\end{equation}
where $\hat{\boldsymbol{m}}(\boldsymbol{y})$ is injected by concatenating with the input $\boldsymbol{d}_t$ along the channel dimension. 

To generate samples from the trained model, the reverse process is iteratively performed starting from a Gaussian noise $\boldsymbol{d}_T \sim \mathcal{N}(0, \mathbf{I})$ as follows:
\begin{equation}\label{eq:sampling}
\mathbf{\boldsymbol{d}}_{t-1} = \frac{1}{\sqrt{\alpha_t}} \left( \mathbf{\boldsymbol{d}}_t - \frac{1 - \alpha_t}{\sqrt{1 - \bar{\alpha}_t}}\boldsymbol{\epsilon}_\theta(\mathbf{\boldsymbol{d}}_t, t, \hat{\boldsymbol{m}}(\boldsymbol{y})) \right) + \sigma_t \boldsymbol{\epsilon}, \quad \boldsymbol{\epsilon} \sim \mathcal{N}(\mathbf{0}, \mathbf{I}).
\end{equation}

This process is repeated until $t = 1$, at which point $\boldsymbol{d}_0$ is taken as the final reconstructed sample. In all experiments, the conditional diffusion model adopts a standard U-Net backbone as the denoising network $\boldsymbol{\epsilon}_\theta$. The coarse-scale condition $\hat{\boldsymbol{m}}(\boldsymbol{y})$ is concatenated with the noisy input along the channel dimension and jointly processed by the U-Net during denoising. Detailed implementation settings of the U-Net backbone used in different experiments are summarized in Appendix.

\subsubsection{Training stage: Mask-cascade training}\label{sec: MCT}

\begin{figure}[H]
    \centering
    \includegraphics[width=0.9\linewidth]{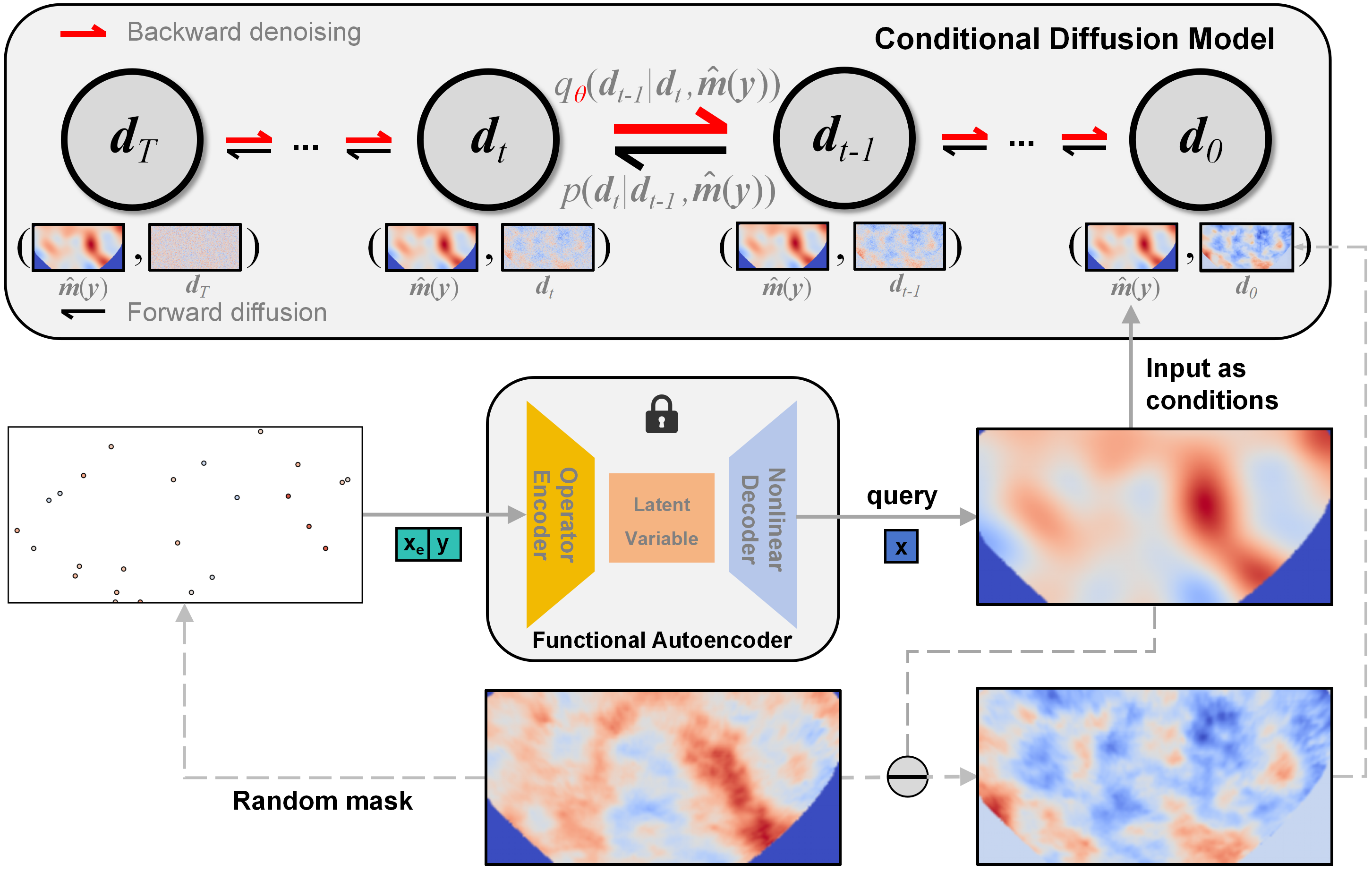}
    \caption{The proposed mask-cascade training (MCT) of the conditional diffusion model, through which the conditional diffusion model learns a coarse-conditioned detail prior $p(\boldsymbol{d}\mid\hat{\boldsymbol{m}}(\boldsymbol{y}))$ without explicitly accessing the corresponding sparse observation $\boldsymbol{y}$ during training.}
    \label{fig:mask-cascade training of CDM_new}
\end{figure}

A key challenge is that the functional autoencoder, when provided with different sparse samplings from the same underlying field, reconstructs a family of slightly different coarse-scale structures $\hat{\boldsymbol{m}}(\boldsymbol{y})$. Under extreme sparsity, this mask-induced variability in the coarse structural estimate is non-negligible. If the conditional diffusion model were trained only on a fixed coarse condition for each field, such variability would be lost, and the learned detail prior could become overly specialized to a narrow set of conditions. To address this issue, we propose a mask-cascade training strategy that explicitly reintroduces this conditioning diversity into diffusion training.

Our strategy explicitly integrates the stochasticity of sparse sampling into training the conditional diffusion model. As illustrated in Fig. \ref{fig:mask-cascade training of CDM_new}, once the functional autoencoder is fully trained and its parameters are frozen, the training of the conditional diffusion model proceeds as follows: in each training step, we randomly generate a sparse mask with fixed mask ratio and apply it to the ground-truth field $\boldsymbol{u}$ to obtain a partial observation $\boldsymbol{y}$. Notably, the mask ratio here is chosen to be significantly lower than that used for masked training the functional autoencoder, so as to explicitly train the conditional diffusion model to operate under extremely sparse observations. This observation is then passed through the trained (fixed) functional autoencoder to reconstruct an approximate coarse-scale structure $\hat{\boldsymbol{m}}(\boldsymbol{y})$. The conditional diffusion model is then conditioned on $\hat{\boldsymbol{m}}(\boldsymbol{y})$, and trained to generate the refined-scale component $\boldsymbol{d} = \boldsymbol{u} - \hat{\boldsymbol{m}}(\boldsymbol{y})$. By repeatedly implementing different masks across training steps, the conditional diffusion model is continuously exposed to a diverse ensemble of approximate coarse-scale structure reconstructions. This effectively augments the training distribution of conditioning inputs without the need for explicitly precomputing or storing all possible variants. As a result, the conditional diffusion model learns to model the conditional distribution $p(\boldsymbol{d}|\boldsymbol{m})$, thereby achieving improved robustness and consistency across varying sparsity patterns.

\subsubsection{Inference stage: Manifold constrained gradient}\label{sec: MCG}

The conditional detail prior $p(\boldsymbol{d}|\boldsymbol{m})$ has been modeled by training the conditional diffusion model with the proposed MCT. According to the formulation in Eq.\eqref{eq:bayes_decomp}, the likelihood $p(\boldsymbol{y}\mid \boldsymbol{d},\hat{\boldsymbol{m}}(\boldsymbol{y}))$ should be modeled to further compress the uncertainty of reconstruction by imposing measurement consistency. As mentioned above, directly modeling the conditional distribution $p(\boldsymbol{d}_{t-1}|\boldsymbol{d}_{t},\hat{\boldsymbol{m}}(\boldsymbol{y}),\boldsymbol{y})$ is impractical due to the immense variability of sparse measurement patterns. We model it during the inference stage with soft conditioning. Specifically, the Manifold Constrained Gradient (MCG) \cite{chung2022improving} is used to inject an additional gradient term to guide the generation towards observed measurements without retraining the model. This gradient refers to the score function in score-based generative models (SGMs) \cite{song2020score}, defined as the gradient of the log-density of the data distribution $\nabla_{\boldsymbol{d}_t} \log p_t(\boldsymbol{d}_t|\hat{\boldsymbol{m}}(\boldsymbol{y}))$. 

Although DDPMs and SGMs were originally proposed under different formulations -- predicting added noise in DDPMs and estimating score functions in SGMs -- it has been shown that they are theoretically equivalent under certain parameterizations \cite{song2020score}. Specifically, when a DDPM is trained to predict the Gaussian noise $\boldsymbol{\epsilon}_{\theta}$ added at each step, the network implicitly learns the score function of the perturbed data distribution $p_t(\boldsymbol{d}_t|\hat{\boldsymbol{m}}(\boldsymbol{y}))$ up to a scaling factor:
\begin{equation}
\nabla_{\boldsymbol{d}_t} \log p_t(\boldsymbol{d}_t|\hat{\boldsymbol{m}}(\boldsymbol{y})) \approx -\frac{1}{\sqrt{1 - \bar{\alpha}_t}}\boldsymbol{\epsilon}_\theta(\boldsymbol{d}_t, t, \hat{\boldsymbol{m}}(\boldsymbol{y})),
\end{equation}

% Therefore, the sampling equation in Eq.\eqref{eq:sampling} can be reformulated as:
% \begin{equation}\label{eq:gradient sampling}
% \mathbf{\boldsymbol{d}}_{t-1} = \frac{1}{\sqrt{\alpha_t}} \left( \mathbf{\boldsymbol{d}}_t + (1 - \alpha_t)\nabla_{\boldsymbol{d}_t} \log p_t(\boldsymbol{d}_t|\hat{\boldsymbol{m}}(\boldsymbol{y})) \right) + \sigma_t \boldsymbol{\epsilon}, \quad \boldsymbol{\epsilon} \sim \mathcal{N}(\mathbf{0}, \mathbf{I}).
% \end{equation}

This connection enables MCG to be directly applied to DDPMs. Specifically, the Bayes rule $p(\boldsymbol{d}_t|\hat{\boldsymbol{m}}(\boldsymbol{y}),\boldsymbol{y}) = \frac{p(\boldsymbol{d}_t| \hat{\boldsymbol{m}}(\boldsymbol{y})) p(\boldsymbol{y}|\boldsymbol{d}_t,\hat{\boldsymbol{m}}(\boldsymbol{y}))}{p(\boldsymbol{y}| \hat{\boldsymbol{m}}(\boldsymbol{y}))}$ leads to:
\begin{equation}\label{eq:Bayes}
    \nabla_{\boldsymbol{d}_t} \log p_t(\boldsymbol{d}_t|\hat{\boldsymbol{m}}(\boldsymbol{y}),\boldsymbol{y})=\nabla_{\boldsymbol{d}_t} \log p_t(\boldsymbol{d}_t|\hat{\boldsymbol{m}}(\boldsymbol{y})) + \nabla_{\boldsymbol{d}_t} \log p_t(\boldsymbol{y} | \boldsymbol{d}_t, \hat{\boldsymbol{m}}(\boldsymbol{y})),
\end{equation}
where $\nabla_{\boldsymbol{d}_t} \log p_t(\boldsymbol{d}_t|\hat{\boldsymbol{m}}(\boldsymbol{y}))$ where \(\nabla_{d_t}\log p_t(d_t|\hat{m}(y))\) is provided by the conditional diffusion model. To approximate the likelihood term $p_t(\boldsymbol{y} | \boldsymbol{d}_t, \hat{\boldsymbol{m}}(\boldsymbol{y}))$, we assume that it follows a Gaussian form. Under this assumption, its gradient with respect to $\boldsymbol{d}_t$ yields:
\begin{equation}\label{eq:project method likelihood}
    \nabla_{\boldsymbol{d}_t} \log p_t(\boldsymbol{y} | \boldsymbol{d}_t, \hat{\boldsymbol{m}}(\boldsymbol{y}))=-\frac{1}{\sigma_c^2}\frac{\partial}{\partial\boldsymbol{d}_t}||\boldsymbol{y}-\mathcal{M}(\hat{\boldsymbol{m}}(\boldsymbol{y})+\boldsymbol{d}_t)||^2_2,
\end{equation}
% , i.e.,
% \begin{equation}
% p_t(\boldsymbol{y}|\boldsymbol{d}_t,\hat{\boldsymbol{m}}(\boldsymbol{y})) \propto 
% \exp\left(-\frac{1}{2\sigma_c^2}\|\boldsymbol{y}-\mathcal{M}(\hat{\boldsymbol{m}}(\boldsymbol{y})+\boldsymbol{d}_t)\|_2^2\right),
% \end{equation}
where $\mathcal{M}(\cdot)$ denotes the mask operator and $\sigma_c^2$ denotes the variance associated with measurement noise.

According to the Bayes rule in Eq. \ref{eq:Bayes}, the measurement consistency can be injected into the trained diffusion model by adding this new gradient into the sampling formulation in Eq. \ref{eq:sampling}. However, it drives the intermediate states at each denoising step to match the observations, which pulls the sample path away from the learned data manifold. Chung et al. \cite{chung2022improving} show that applying measurement guidance to the Tweedie denoised estimate \cite{robbins1992empirical} provides a more stable conditioning target for diffusion sampling, both empirically and theoretically. As a result, Tweedie’s formula is introduced here to impose the measurement consistency while improving the generation accuracy. In the case of Gaussian noise, a classic result of Tweedie’s formula tells us that one can achieve the denoised result by computing the posterior expectation:
\begin{equation}
    \mathbb{E}[\boldsymbol{d}|\tilde{\boldsymbol{d}}]=\tilde{\boldsymbol{d}}+\sigma^2\nabla_{\tilde{\boldsymbol{d}}}log\ p(\tilde{\boldsymbol{d}}),
\end{equation}
where the noise is modeled by $\tilde{\boldsymbol{d}} \sim \mathcal{N}(\boldsymbol{d}, \sigma^2\mathbf{I})$. Considering the conditional diffusion model where the forward step is modeled as $\boldsymbol{d}_t\sim\mathcal{N}\left(\sqrt{\bar{\alpha}_t} \, \boldsymbol{d}_0, \, (1 - \bar{\alpha}_t) \, \mathbf{I} \right)$, the Tweedie’s formula can be rewritten as:
\begin{equation}\label{eq:Tweedie's formula}
    \mathbb{E}[\boldsymbol{d}_0|\boldsymbol{d}_t,\hat{\boldsymbol{m}}(\boldsymbol{y})]=\frac{1}{\sqrt{\bar{\alpha}_t}}\big(\boldsymbol{d}_t+(1 - \bar{\alpha}_t)\nabla_{\boldsymbol{d}_t} \log p_t(\boldsymbol{d}_t|\hat{\boldsymbol{m}}(\boldsymbol{y}))\big)
\end{equation}

By replacing $\boldsymbol{d}_t$ in Eq.~\eqref{eq:project method likelihood} with the posterior expectation computed by Tweedie's formula $\hat{\boldsymbol{d}}_0(\boldsymbol{d}_t)=\mathbb{E}[\boldsymbol{d}_0|\boldsymbol{d}_t,\hat{\boldsymbol{m}}(\boldsymbol{y})]$, the sampling formula for the trained conditional diffusion model is given as:
\begin{equation}\label{eq:mcg sample}
\boldsymbol{d}_{t-1} = \frac{1}{\sqrt{\alpha_t}} \left( \mathbf{\boldsymbol{d}}_t + (1 - \alpha_t)\big(\nabla_{\boldsymbol{d}_t} \log p_t(\boldsymbol{d}_t|\hat{\boldsymbol{m}}(\boldsymbol{y}))-\frac{1}{\sigma_c^2}\frac{\partial}{\partial\boldsymbol{d}_t}||\boldsymbol{y}-\mathcal{M}(\hat{\boldsymbol{m}}(\boldsymbol{y})+\hat{\boldsymbol{d}}_0(\boldsymbol{d}_t))||^2_2\big)\right) + \sigma_t \boldsymbol{\epsilon}, \quad \boldsymbol{\epsilon} \sim \mathcal{N}(\mathbf{0}, \mathbf{I}),
\end{equation}
where $-\frac{1}{\sigma_c^2}\frac{\partial}{\partial\boldsymbol{d}_t}||\boldsymbol{y}-\mathcal{M}(\hat{\boldsymbol{m}}(\boldsymbol{y})+\hat{\boldsymbol{d}}_0(\boldsymbol{d}_t))||^2_2$ is the manifold constrained gradient that enforces consistency between the predicted clean sample and the observations. It has been proven that manifold constrained gradient is the projection of Eq.~\eqref{eq:project method likelihood} onto the data manifold \cite{chung2022improving}, thereby keeping the trajectory on the data manifold and alleviating sample degradation.

\begin{figure}[H]
    \centering
    \includegraphics[width=1.\linewidth]{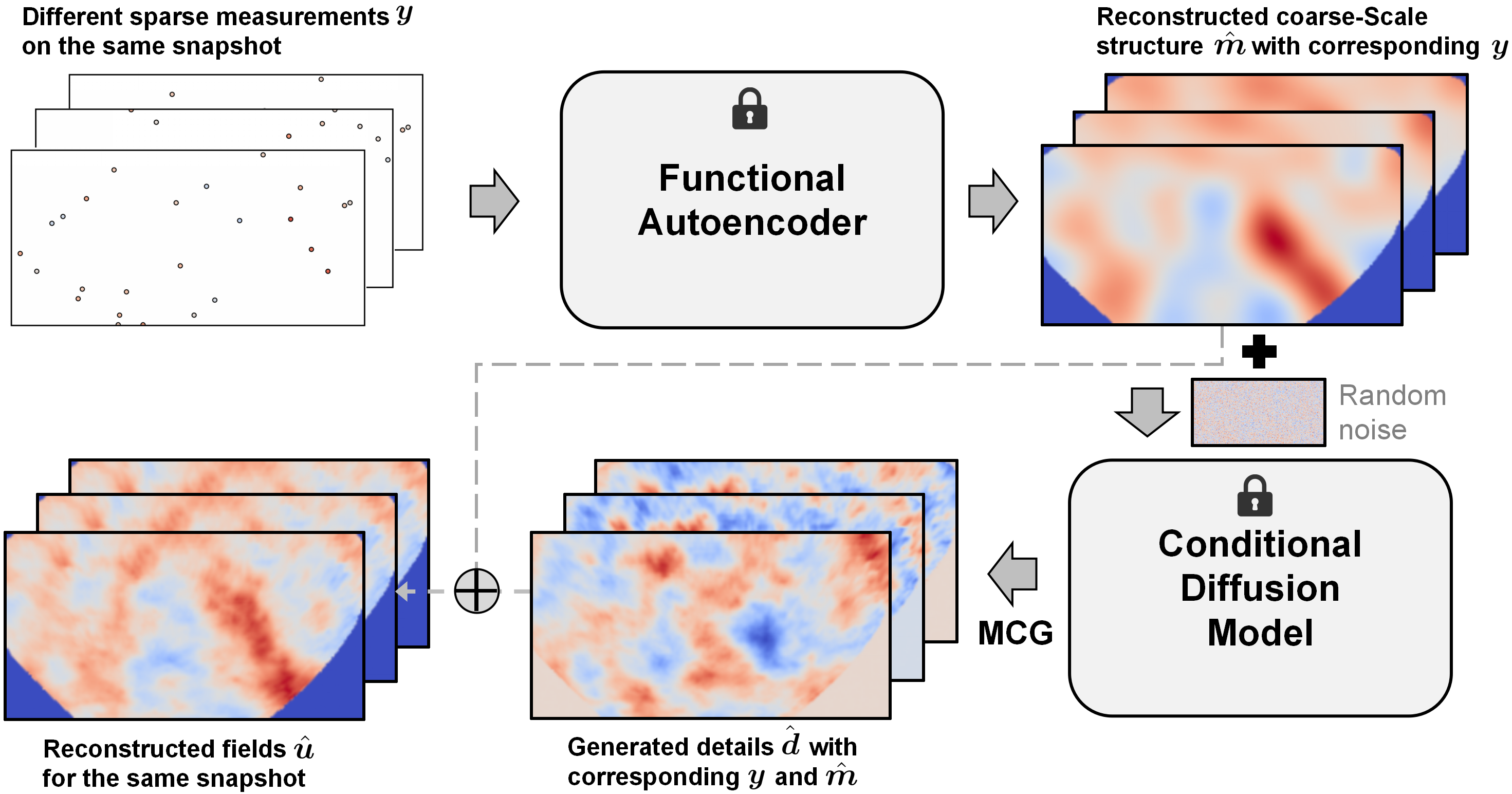}
    \caption{The inference scheme of Cascaded Sensing. Given different sparse measurements of the same target physical field, the framework samples from the conditional distribution
$p(\boldsymbol{u}|\boldsymbol{y}) \approx p(\boldsymbol{d}|\hat{\boldsymbol{m}}(\boldsymbol{y}),\boldsymbol{y})$,
producing multiple reconstruction samples. Measurement consistency is imposed with Manifold Constrained Gradient (MCG) during diffusion sampling, without retraining the model.}
    \label{fig:inference of Cas-Sensing}
\end{figure}

As shown in Fig.~\ref{fig:inference of Cas-Sensing}, the inference process of autoencoder-diffusion cascade proceeds as follows. Given arbitrary sparse observations $\boldsymbol{y}$, the functional autoencoder first reconstructs the coarse-scale structure $\hat{\boldsymbol{m}}(\boldsymbol{y})$. This reconstructed structure is then provided as a condition to the conditional diffusion model, which generates the refined-scale details $\hat{\boldsymbol{d}}$ through the manifold constrained gradient-enhanced sampling scheme in Eq. \eqref{eq:mcg sample}. Finally, the full-field reconstruction is obtained by combining the two components $\hat{\boldsymbol{u}} = \hat{\boldsymbol{m}}(\boldsymbol{y}) + \hat{\boldsymbol{d}}$. It is worth noting that, during inference, Cas-Sensing can still achieve reliable reconstruction even when the sampling rate of sparse observations differs significantly from the rates used in mask-cascade training.

\section{Results}\label{sec:results}

\subsection{Reconstructing sea surface wave height fields with stereo data}

\begin{figure}[H]
\centering 
\includegraphics[width=0.7\textwidth]{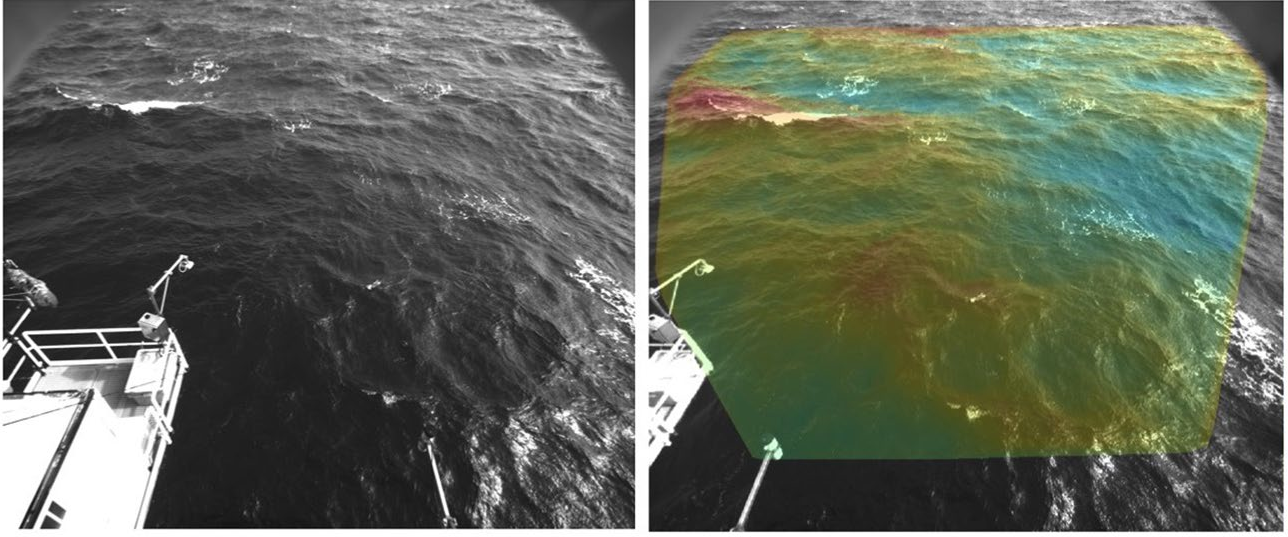}
\caption{An example of stereo-image pair (from the Acqua Alta tower in the northern Adriatic Sea, Italy) and 3-D wave field on top of the right image (the colors cale is proportional to the sea surface height) \cite{guimaraes2020data}.}
\label{fig:WASS example}
\end{figure}

In this subsection, we demonstrate Cas-Sensing on stereo image data of sea surface wave height fields for reconstruction from sparse and noisy measurements, highlighting its utility in a complex real-world engineering setting. Stereo imaging measurement of the sea surface height is based on single snapshots or time records captured by a pair of synchronized and calibrated cameras. An example of a stereo-image pair processed by the Wave Acquisition Stereo System (WASS) is presented in Fig.~\ref{fig:WASS example}. The dataset used in this study is the Acqua Alta stereo dataset, which contains 8000 image frames recording a half-hour sea-state evolution with an imaging system installed on the north-east side of the Acqua Alta oceanographic research tower. The wave fields are reshaped to a resolution of $256 \times 128$.

\begin{figure}[H]
\centering 
\includegraphics[width=1.\textwidth]{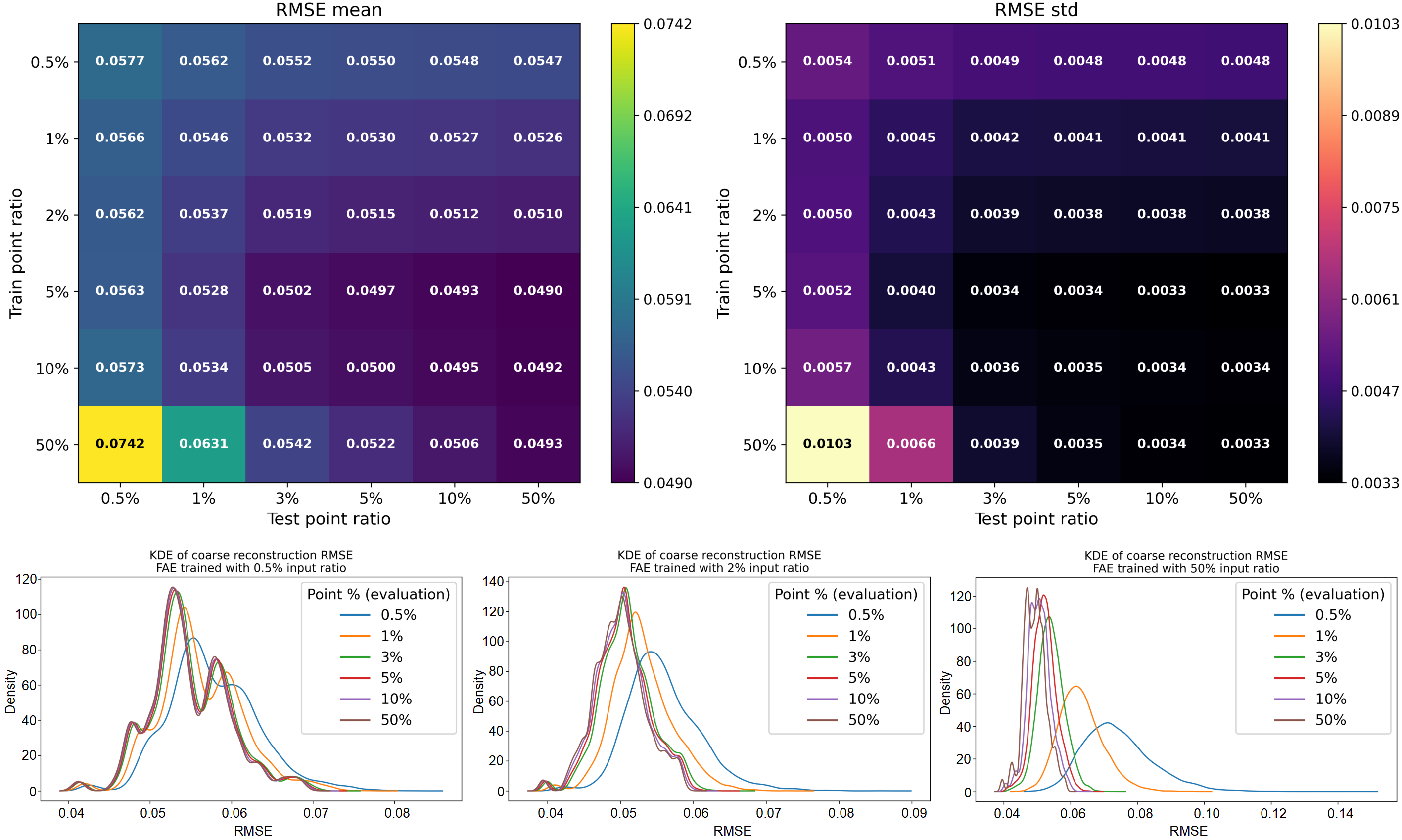}
\caption{Comparison of coarse-scale reconstruction performance for functional autoencoders trained with different input point ratios. The heat maps report the mean and standard deviation of RMSE over 100 randomly selected test samples and 100 random masks per sample, and the kernel density plots show representative RMSE distributions for selected training ratios across different testing point ratios.}
\label{fig:sea surface FAE ratio comparision}
\end{figure}

Before evaluating the full Cas-Sensing framework, we first assess whether the functional autoencoder can stably recover the coarse-scale structure $\boldsymbol{m}$ from sparse observations under varying sensing conditions. This step is essential because the advantage of Cas-Sensing over purely soft-conditioned diffusion depends on whether the functional autoencoder can provide an accurate and robust structural condition for the downstream conditional diffusion model. To this end, after normalizing the original dataset with min-max normalization, we train a functional autoencoder with latent dimension 128 on 7000 image frames and reserve the remaining 1000 frames for testing. To examine robustness under different observation sparsity levels, we train separate functional autoencoders using input point ratios of $0.5\%, 1\%, 2\%, 5\%, 10\%$, and $50\%$, respectively. Each model is trained to reconstruct the coarse-scale field from sparse observations generated at its corresponding input ratio. We then evaluate every trained model under six testing point ratios, namely $0.5\%, 1\%, 3\%, 5\%, 10\%$, and $50\%$, in order to systematically assess how the training sparsity level affects coarse-scale reconstruction across different sensing conditions. Specifically, for each testing ratio, we randomly select 100 samples from the test set, and for each sample we generate 100 distinct random masks at the corresponding sparsity level. Each masked input is passed through the encoder-decoder pipeline to reconstruct the coarse-scale field on the full grid, and the resulting root mean square error (RMSE) values over all samples and masks are aggregated to form the kernel density estimates and summary statistics shown in Fig. \ref{fig:sea surface FAE ratio comparision}.

Several clear trends can be observed from Fig. \ref{fig:sea surface FAE ratio comparision}. First, except for the extremely dense training case, both the mean and the standard deviation of the reconstruction RMSE generally decrease as the testing point ratio increases, indicating that denser observations provide more reliable information for recovering the coarse-scale structure. Second, the training sparsity level has a non-monotonic effect on generalization. When the training input is excessively sparse, such as $0.5\%$, the model performs poorly across all testing ratios, suggesting that the available observations are insufficient for the functional autoencoder to learn a stable sparse-to-coarse reconstruction law. In contrast, when the training input is excessively dense, such as $50\%$, the model performs well only when the testing observations are also dense, but deteriorates markedly under sparse testing conditions. This indicates that the model is no longer trained to infer coarse-scale structures from highly incomplete inputs; instead, the task becomes closer to direct field reconstruction from relatively rich observations, which weakens robustness under sparse sensing at test time. Between these two extremes, models trained with moderately sparse inputs, particularly around $2\%$ to $5\%$, achieve the best trade-off between accuracy and robustness: they yield the lowest RMSE statistics over most testing ratios and exhibit consistently concentrated error distributions in the kernel density plots. These results show that a reliable hard condition for the downstream conditional diffusion model is obtained not by maximizing the training observation ratio, but by training the functional autoencoder under sufficiently sparse yet still informative sensing conditions. Given our focus on reconstruction under extreme sparsity, the functional autoencoder trained with a $2\%$ input ratio is used in the subsequent experiments to provide the structural-anchor condition for conditional diffusion training and to generate the corresponding refined-scale details.

In addition, to examine robustness to observation noise, we perturb the sparse measurements with additive Gaussian noise at different relative noise levels and evaluate whether the reconstructed coarse-scale structure remains stable under extreme sparsity. Specifically, for each fixed testing point ratio, we randomly select 100 samples from the test set and generate 100 distinct random masks for each sample. Noise is added to the sparse observed values, with the noise standard deviation defined as a prescribed fraction of the standard deviation of the corresponding sparse measurements. The resulting RMSE values are aggregated to form kernel density estimates under different noise levels. These tests are designed to verify whether the functional autoencoder can maintain reliable reconstruction of $\boldsymbol{m}$ under varying noise conditions, which is another key prerequisite for the robustness of the subsequent hard-conditioned diffusion refinement.

\begin{figure}[H]
\centering 
\includegraphics[width=1.\textwidth]{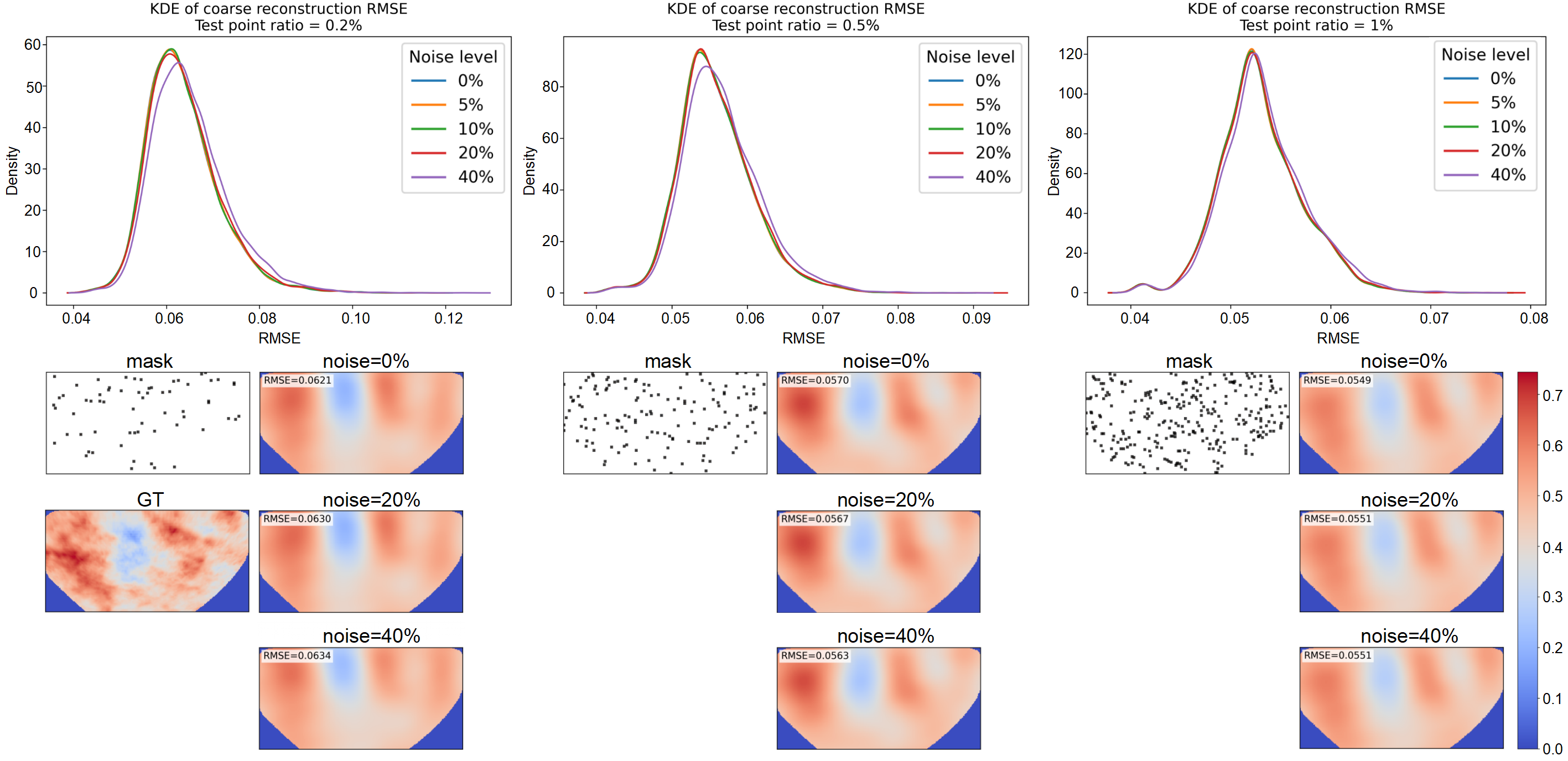}
\caption{Robustness of the functional autoencoder to observation noise under extreme sparsity. For each fixed testing point ratio, the kernel density estimates are computed from RMSE values aggregated over 100 randomly selected test samples and 100 random masks per sample under different relative Gaussian noise levels.}
\label{fig:sea surface FAE noise comparision}
\end{figure}

Figure \ref{fig:sea surface FAE noise comparision} further evaluates the robustness of the functional autoencoder to observation noise under extreme sparsity. For each fixed testing point ratio, the RMSE distributions remain nearly unchanged from $0\%$ to $20\%$ noise, indicating that the coarse-scale reconstruction is largely insensitive to mild-to-moderate perturbations in the sparse measurements. A more visible degradation appears only at the strongest noise level of $40\%$, where the distributions shift slightly toward larger errors and become moderately broader. The effect of noise is also somewhat stronger at lower testing point ratios, which is consistent with the increased influence of each observed point in more severely underdetermined settings. However, even at $0.2\%$ observations, the overall variation in the error distribution remains limited. This confirms that the functional autoencoder can maintain a stable reconstruction of $\boldsymbol{m}$ across a broad range of noise levels, thereby providing a reliable hard condition for the downstream conditional diffusion model.

\begin{figure}[H]
\centering
\includegraphics[width=0.9\textwidth]{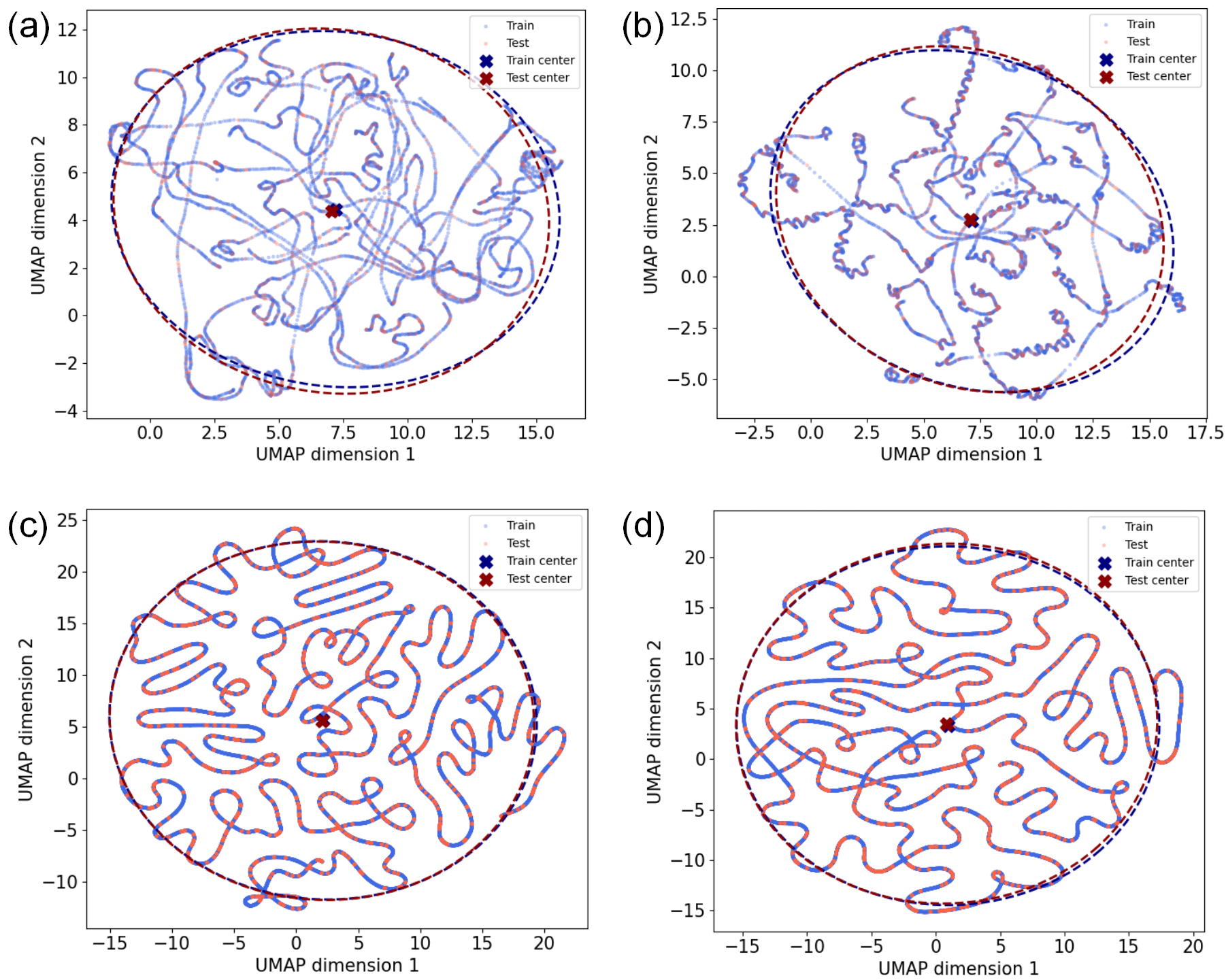}
\caption{Comparison of data manifolds visualized by PCA followed by UMAP. (a) Original full-field data manifold. (b) Refined-scale detail data manifold obtained after removing the ``perfect'' coarse-scale structure, where the coarse-scale structure is produced by feeding the complete observations into the pretrained functional autoencoder. (c) Refined-scale detail data manifold induced by mask-cascade training with $0.5\%$ input ratio. (d) Refined-scale detail data manifold induced by mask-cascade training with $0.2\%$ input ratio.}
\label{fig:sea surface manifold comparison}
\end{figure}

Before training the conditional diffusion model with mask-cascade training (MCT), we further examine the geometry of the generative targets in Fig.~\ref{fig:sea surface manifold comparison} by comparing the manifolds of the original full-field data, the refined-scale detail data obtained after removing the ``perfect'' coarse-scale structure, and the refined-scale detail data induced by MCT under different sparse input ratios. Here, the ``perfect'' coarse-scale structure is defined as the output of the pretrained functional autoencoder when all data points are provided as input. This reference detail manifold characterizes the refined-scale target produced by the coarse-to-detail decomposition itself. In MCT, after fixing the pretrained functional autoencoder, we generate coarse-scale structures from extremely sparse observations using input ratios of $0.5\%$ and $0.2\%$, corresponding to mask ratios of $99.5\%$ and $99.8\%$, respectively. These highly sparse masking strategies are intentionally adopted to induce diverse coarse-scale structural anchors, which in turn produce a family of refined-scale details for conditional diffusion training. For manifold visualization, each sample is propagated through the fixed functional autoencoder ten times using different random masks, and the resulting refined-scale details are embedded by PCA followed by UMAP. As shown in Fig.~\ref{fig:sea surface manifold comparison}, even when the input ratio is only $0.2\%$, the refined-scale detail distributions remain organized on structured low-dimensional manifolds. This indicates that MCT does not destabilize the generative target of the conditional diffusion model. Instead, it partially reintroduces the uncertainty compressed by deterministic coarse reconstruction through mask-induced condition variability, while preserving the learnability of the detail distribution. As a result, the conditional diffusion model is exposed to a broader yet still well-structured family of plausible detail patterns, which improves its robustness under varying sparse observation conditions.

\begin{figure}[H]
\centering 
\subfigure[Comparison between Cas-Sensing and Uncond+MCG under $0.1\%$ noise-free observations across different observation likelihood weights. The left panel reports the full-field RMSE and observation error (ObsErr) as functions of the MCG weight. The right panel shows the modes at $\frac{1}{\sigma_c^2}=10^9$ of both methods. The bottom panels further present the distributions of RMSE and ObsErr over 100 generated samples at $\frac{1}{\sigma_c^2}=10^9$.]{
\label{fig:sea_surface_RMSE_ObsErr_uncond_comparison}
\includegraphics[width=1\textwidth]{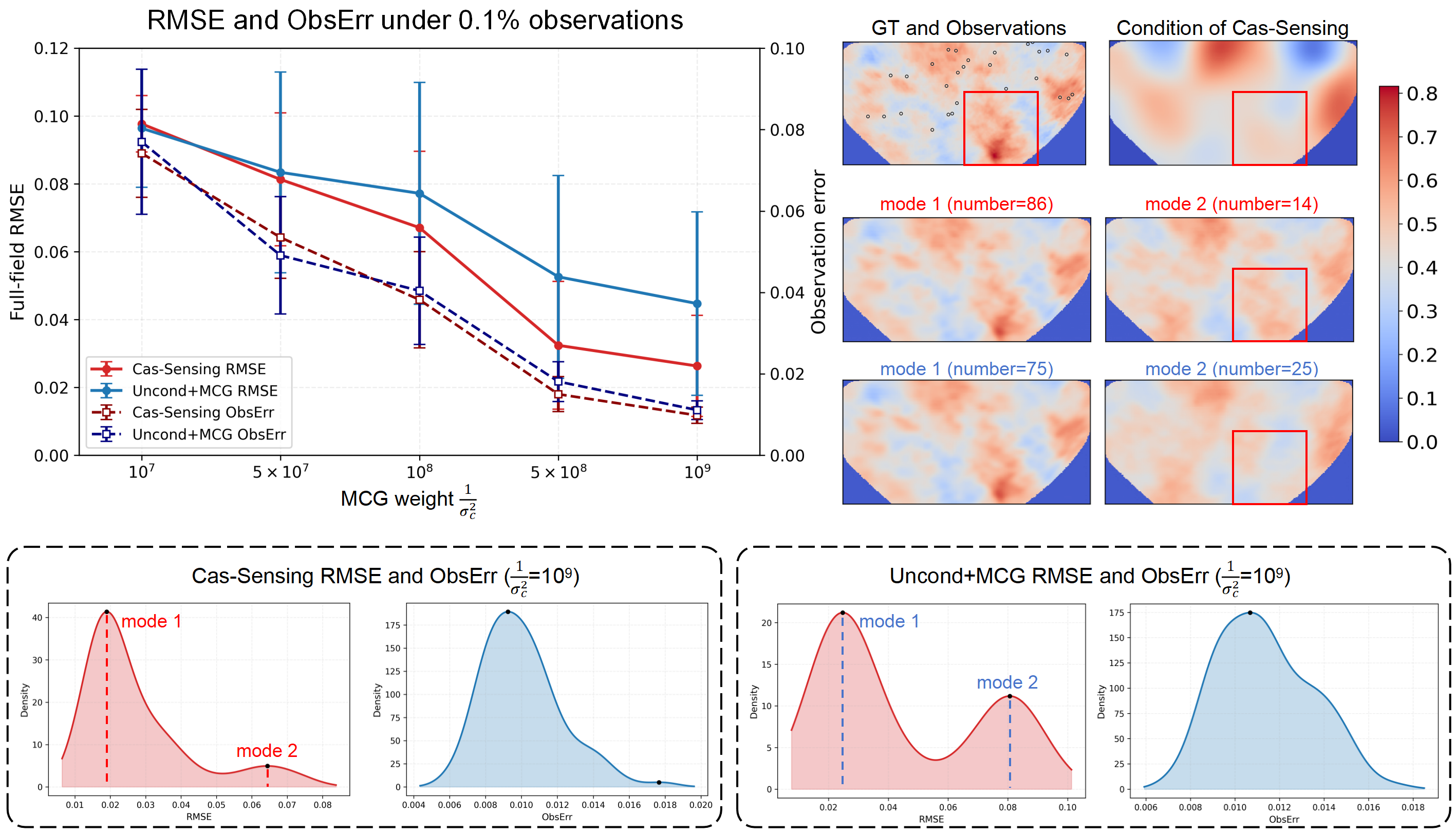}}
\subfigure[Comparison of reverse diffusion trajectories between Cas-Sensing and Uncond+MCG with likelihood weight $\frac{1}{\sigma_c^2}=10^9$, visualized using projection onto the first principal component (PC1).]{
\label{fig:sea_surface_tracjectories_uncond_comparison}
\includegraphics[width=1\textwidth]{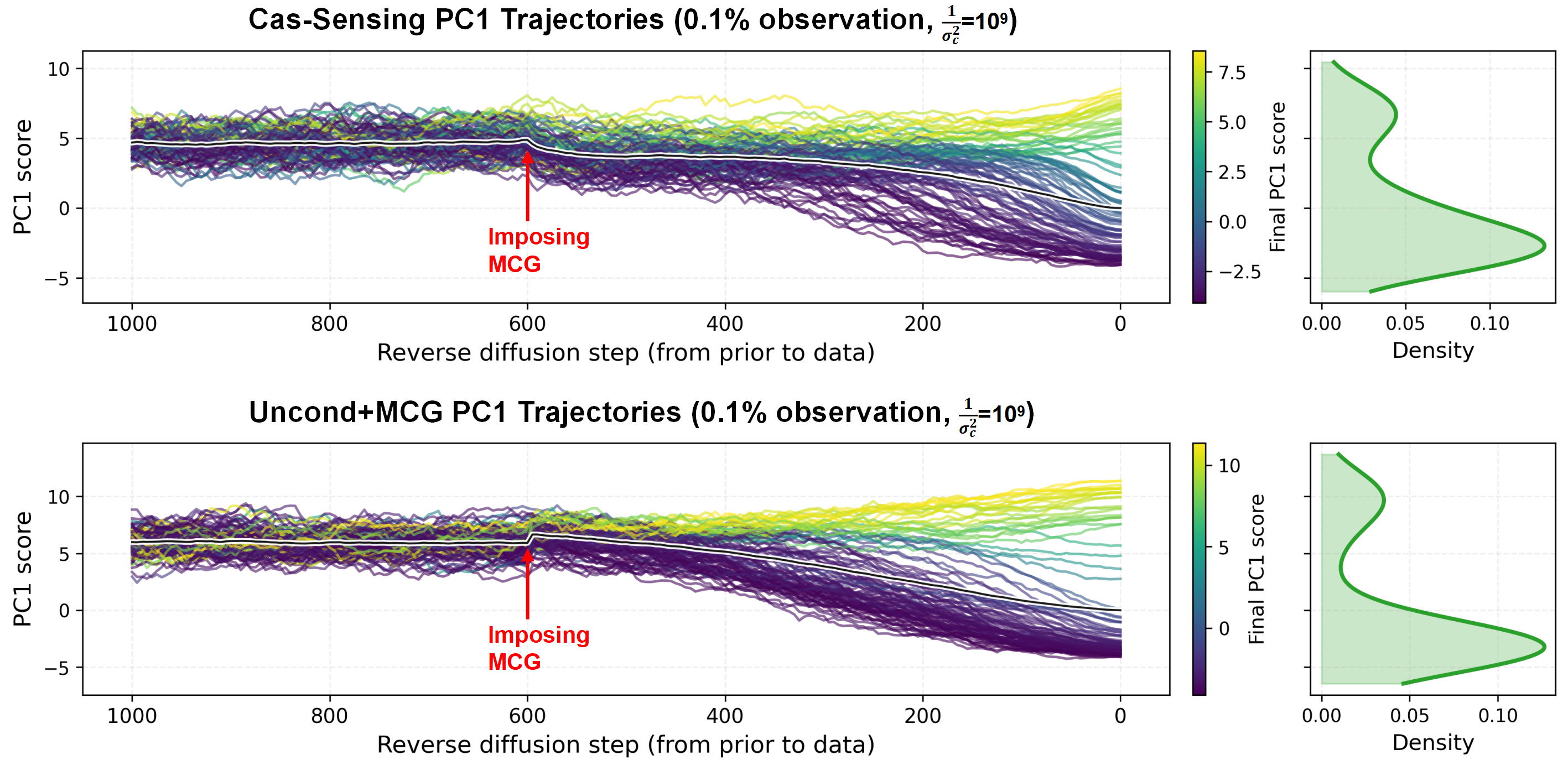}}
\caption{Comparison between Cas-Sensing and Uncond+MCG under extremely sparse noise-free observations with varying likelihood weights, highlighting both statistical performance and underlying sampling dynamics.}
\label{fig:sea_surface_cas-sensing_uncond_mcg_comparison}
\end{figure}

We evaluate the effectiveness of the proposed Cas-Sensing framework by comparing it with an unconditional diffusion model combined with manifold-constrained guidance (Uncond+MCG) under extreme sparsity. In Cas-Sensing, the conditional diffusion model (CDM) is trained using the proposed mask-cascade training (MCT) strategy, where $0.2\%$ of spatial points are randomly sampled and passed through a pre-trained functional autoencoder (FAE) to produce a coarse-scale structural estimate. This estimate serves as a conditioning input to the CDM, which is then trained to model the corresponding refined-scale residual distribution. The CDM is trained for 1000 epochs. In contrast, the unconditional diffusion model is trained directly on the original full-field training data for 1000 epochs without any structural conditioning. At test time, both methods are evaluated by performing 100 generations per sample with $0.1\%$ noise-free observations, rather than the $0.2\%$ observation ratio used in MCT, to demonstrate the generalization ability of Cas-Sensing across different sensor configurations, and observation consistency is enforced through MCG with varying likelihood weights, as shown in Fig.~\ref{fig:sea_surface_cas-sensing_uncond_mcg_comparison}.

Fig.~\ref{fig:sea_surface_cas-sensing_uncond_mcg_comparison} compares Cas-Sensing and Uncond+MCG under extremely sparse ($0.1\%$) noise-free observations from both statistical and dynamical perspectives. Both methods require sufficiently large likelihood weights to enforce observation consistency, and at high weights their observation errors (ObsErr) become comparable. However, their full-field reconstruction performance differs markedly: Cas-Sensing achieves consistently lower RMSE, with the gap widening as the guidance strength increases. This shows that accurate reconstruction under extreme sparsity is not determined by observation fitting alone. The difference becomes more apparent when examining the posterior structure at $\frac{1}{\sigma_c^2}=10^9$ ($\frac{1}{\sigma_c^2}=2\times10^9$ leads to numerical failure, with the generation process producing NaN values). While both methods produce concentrated ObsErr distributions, their RMSE distributions remain distinct. Uncond+MCG exhibits two well-separated modes with comparable proportions, indicating that substantially different full-field solutions remain consistent with the sparse observations. In contrast, Cas-Sensing also exhibits multimodality, but with a dominant accurate mode and a much smaller secondary mode, and the two modes are structurally closer. This suggests that the uncertainty retained by Cas-Sensing is more localized and constrained. The mode-mean reconstructions further reveal the origin of this difference. In regions without observations, Uncond+MCG admits alternative solutions that are consistent with both the observations and the learned prior but fail to recover key structures. In contrast, the coarse structural condition inferred by the functional autoencoder already predicts plausible large-scale patterns in these regions. As a result, even secondary modes in Cas-Sensing preserve the main structural features.

The reverse diffusion trajectories provide a complementary perspective. After MCG is imposed, Uncond+MCG trajectories split into distinct branches, reflecting competition between multiple observation-consistent solutions. In contrast, Cas-Sensing trajectories remain tightly clustered around the coarse structural anchor, indicating that MCG primarily acts as a local refinement rather than a global mode-selection mechanism. Overall, these results show that the advantage of Cas-Sensing does not lie in stronger observation fitting, but in its ability to restrict the solution space through deterministic coarse-scale structural inference, thereby stabilizing posterior inference under extreme sparsity.

\begin{figure}[H]
\centering 
\subfigure[Performance comparison between Cas-Sensing and Uncond+MCG under varying Gaussian observation noise levels at $0.1\%$ sparsity.]{
\label{fig:sea_surface_noise_comparison_line}
\includegraphics[width=1.\textwidth]{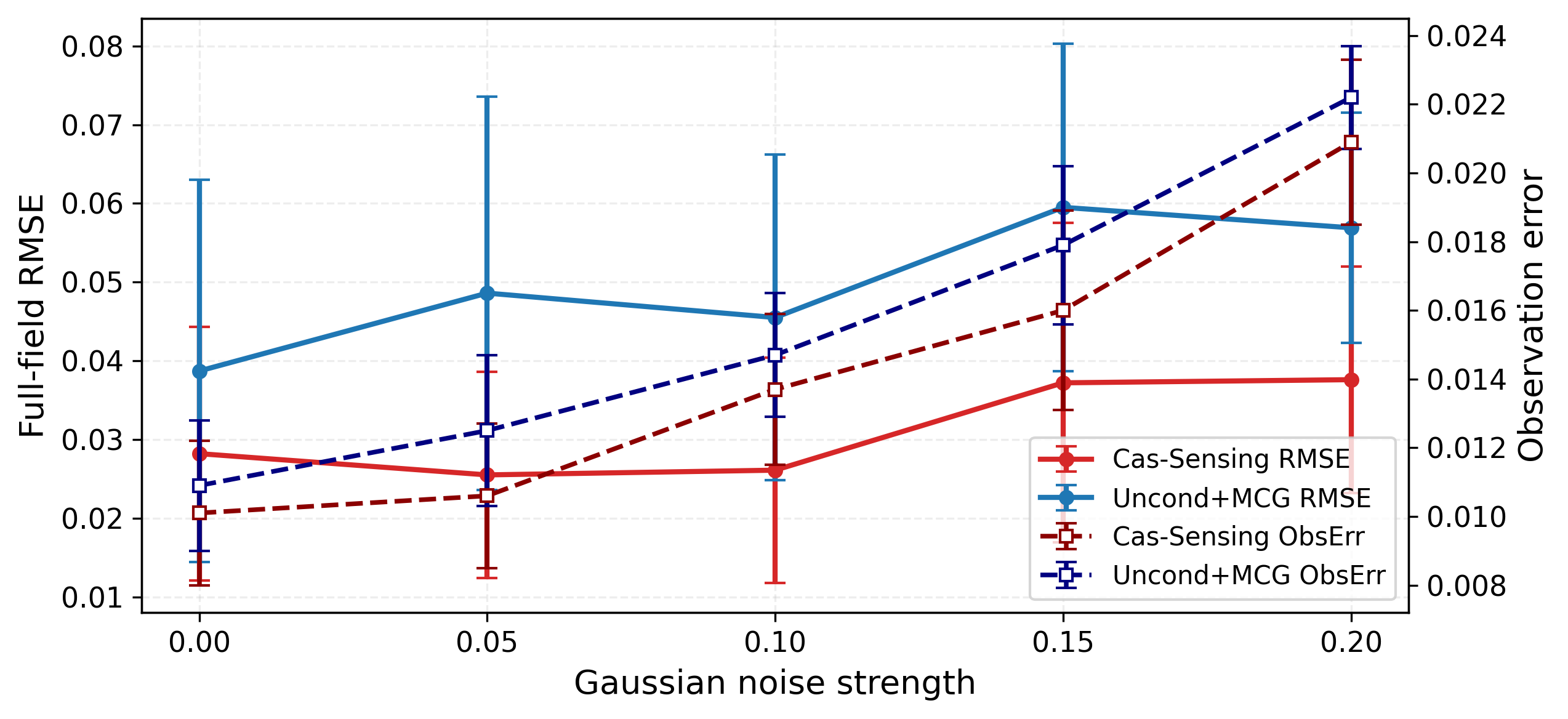}}
\subfigure[KDE and mode comparison of Cas-Sensing and Uncond+MCG under varying Gaussian observation noise levels at $0.1\%$ sparsity.]{
\label{fig:sea_surface_noise_comparison}
\includegraphics[width=1\textwidth]{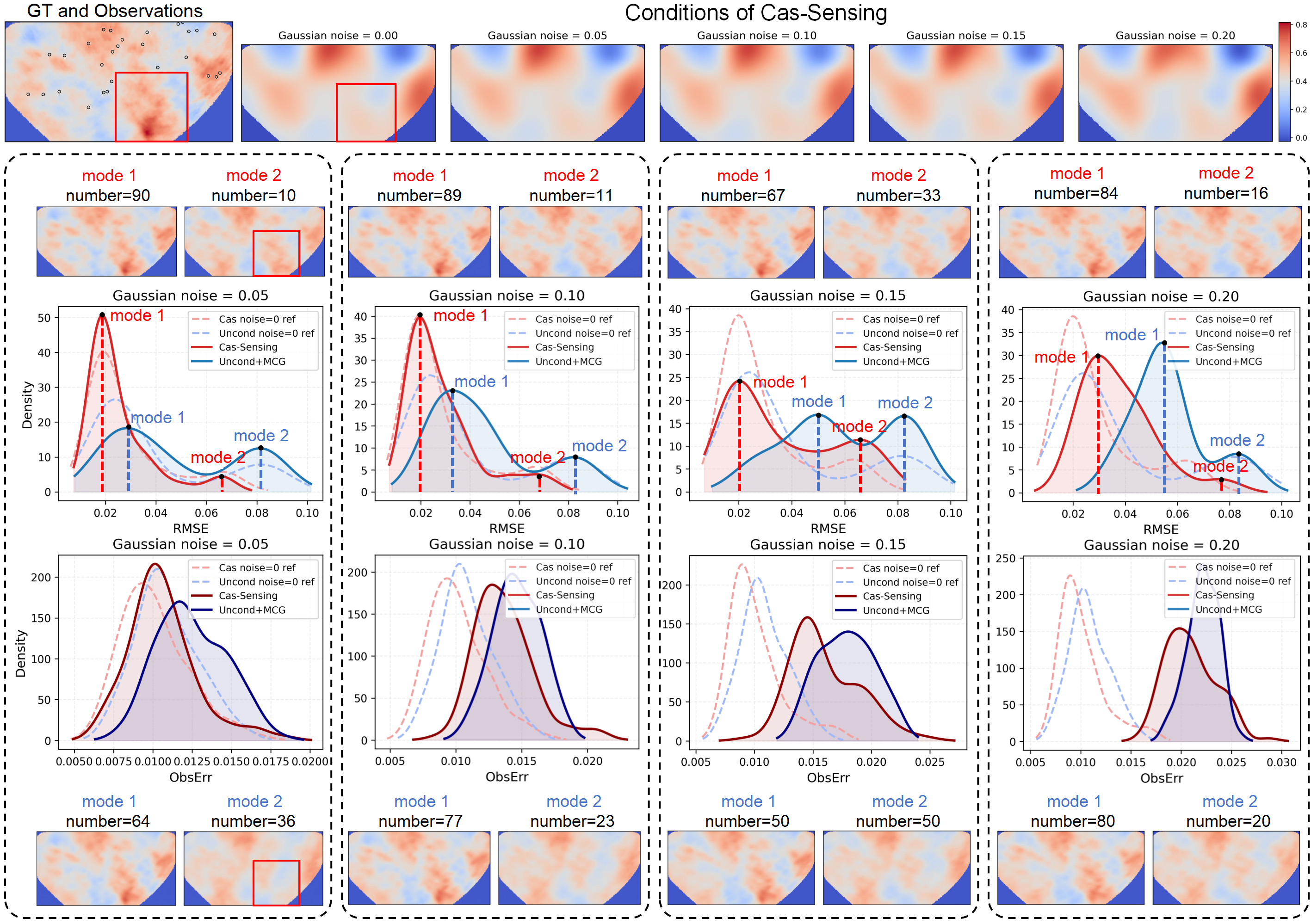}}
\caption{Comparison results of Cas-Sensing and Uncond+MCG under varying Gaussian observation noise levels at $0.1\%$ sparsity.}
\label{fig:sea_surface_noise_comparison_total}
\end{figure}

We next evaluate robustness to observation noise under extreme sparsity ($0.1\%$) with a fixed likelihood weight, as shown in Fig.~\ref{fig:sea_surface_noise_comparison_total}. As noise increases, both methods exhibit similar growth in ObsErr, indicating comparable capability in fitting noisy observations when strong guidance is applied. However, their RMSE behaviors differ substantially: Cas-Sensing consistently achieves lower RMSE, and the gap becomes more pronounced at higher noise levels. Although both methods exhibit multimodal RMSE distributions, their stability differs. For Uncond+MCG, the relative proportions of modes vary significantly with noise, indicating that the inferred posterior is highly sensitive to perturbations in observations. In contrast, Cas-Sensing maintains a stable dominant mode across noise levels, suggesting a more robust posterior approximation. This difference can be attributed to the coarse structural condition. The functional autoencoder consistently aligns with the dominant mode of the posterior, effectively acting as a MAP-like estimator of the large-scale structure. While this mapping discards less probable modes, it anchors the reconstruction around the most likely configuration, thereby stabilizing generation under both extreme sparsity and noise. Importantly, although noise affects both methods through observation guidance, its impact on Cas-Sensing is significantly weaker. The coarse-scale conditions inferred by the autoencoder remain largely unchanged across noise levels, fixing the dominant structural degrees of freedom before the generative stage. As a result, noise-induced uncertainty has limited influence on the final reconstruction. These results show that the robustness of Cas-Sensing arises from stabilizing the posterior structure rather than improving observation fitting.

\begin{figure}[H]
\centering
\includegraphics[width=1.\textwidth]{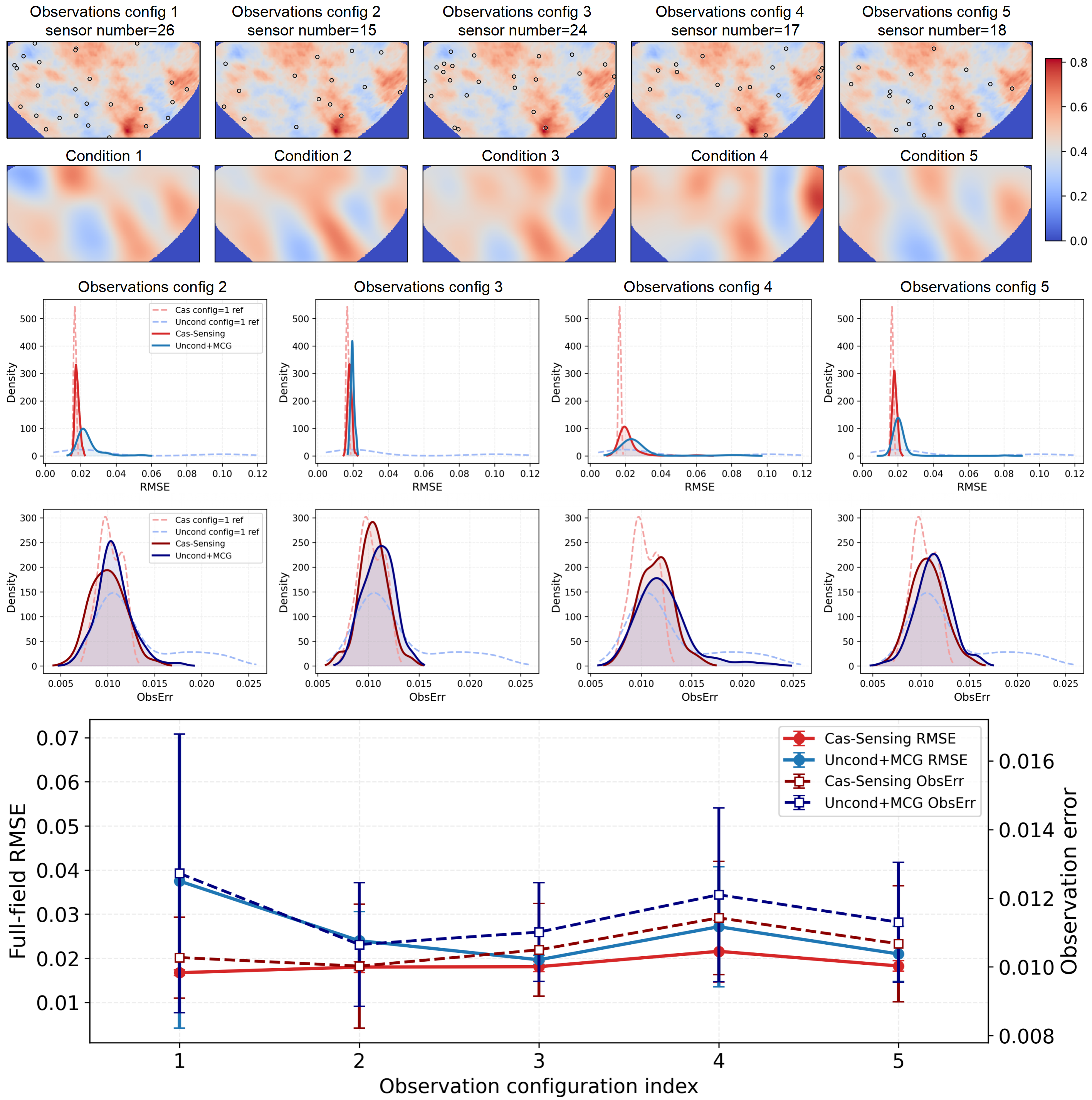}
\caption{Comparison of Cas-Sensing and Uncond+MCG under varying observation configurations (sensor locations and numbers) below $0.1\%$ sparsity without noise and fixed likelihood weight ($1/\sigma_c^2 = 10^9$).}
\label{fig:sea_surface_obs_config_comparison}
\end{figure}

We further examine robustness under varying observation configurations, including different sensor locations and sensor numbers, at extreme sparsity, as shown in Fig.~\ref{fig:sea_surface_obs_config_comparison}. The results reveal that Uncond+MCG is sensitive to both the effective MCG guidance strength and the sensor configuration. In configurations 1 and 4, the RMSE and ObsErr KDEs of the two methods show nearly proportional differences: Uncond+MCG has larger ObsErr and correspondingly larger RMSE. This indicates that, under the fixed likelihood weight used here, MCG does not sufficiently minimize the observation error for Uncond+MCG in these configurations, suggesting that different sensor layouts may require different guidance weights to enforce observation consistency. In configurations 2, 3, and 5, the ObsErr KDEs of Cas-Sensing and Uncond+MCG are comparable, indicating similar observation fitting. However, their full-field RMSE behaviors differ. In configuration 3, the RMSE distributions are also similar, showing that Uncond+MCG can succeed for favorable sensor layouts. In configurations 2 and 5, however, Cas-Sensing retains concentrated RMSE distributions with lower means, whereas Uncond+MCG exhibits much broader RMSE distributions. This indicates that even when sparse observations are fitted similarly, Uncond+MCG remains more sensitive to the sensor layout and may generate globally inaccurate but observation-consistent fields. Overall, these results show that the instability of Uncond+MCG comes from two sources: sensitivity to the MCG weight and sensitivity to the observation operator. Cas-Sensing mitigates both effects by first constraining the dominant large-scale structure through the FAE condition, so that MCG acts mainly as local refinement rather than global mode selection. Consequently, Cas-Sensing achieves more stable reconstruction across different sensor configurations under extreme sparsity.

We further investigate the roles of Mask-Cascade Training (MCT) and Manifold-Constrained Gradient (MCG) through ablation. Without MCG, both variants exhibit large ObsErr, indicating that samples are not consistent with observations and are dominated by the learned prior. When MCG is introduced, the behaviors diverge. For the model without MCT, RMSE distributions remain broad and largely unchanged despite reduced ObsErr. This indicates that observation guidance alone cannot recover accurate solutions, as the correct solutions are not well represented in the learned distribution. The similarity between ``w/o MCT w/o MCG'' and ``w/o MCT with MCG'' confirms that MCG can only project samples within an incorrect solution space. In contrast, for the model trained with MCT, adding MCG immediately collapses both RMSE and ObsErr distributions to a sharp and accurate peak. This indicates that the correct solutions are already contained in the learned distribution, and MCG acts primarily as a selection and refinement mechanism. These results highlight the complementary roles of the two components: MCT expands and stabilizes the support of the conditional distribution, while MCG enforces observation consistency. Without proper support coverage, observation guidance alone is insufficient under distribution shift.

\begin{figure}[H]
\centering
\includegraphics[width=1.\textwidth]{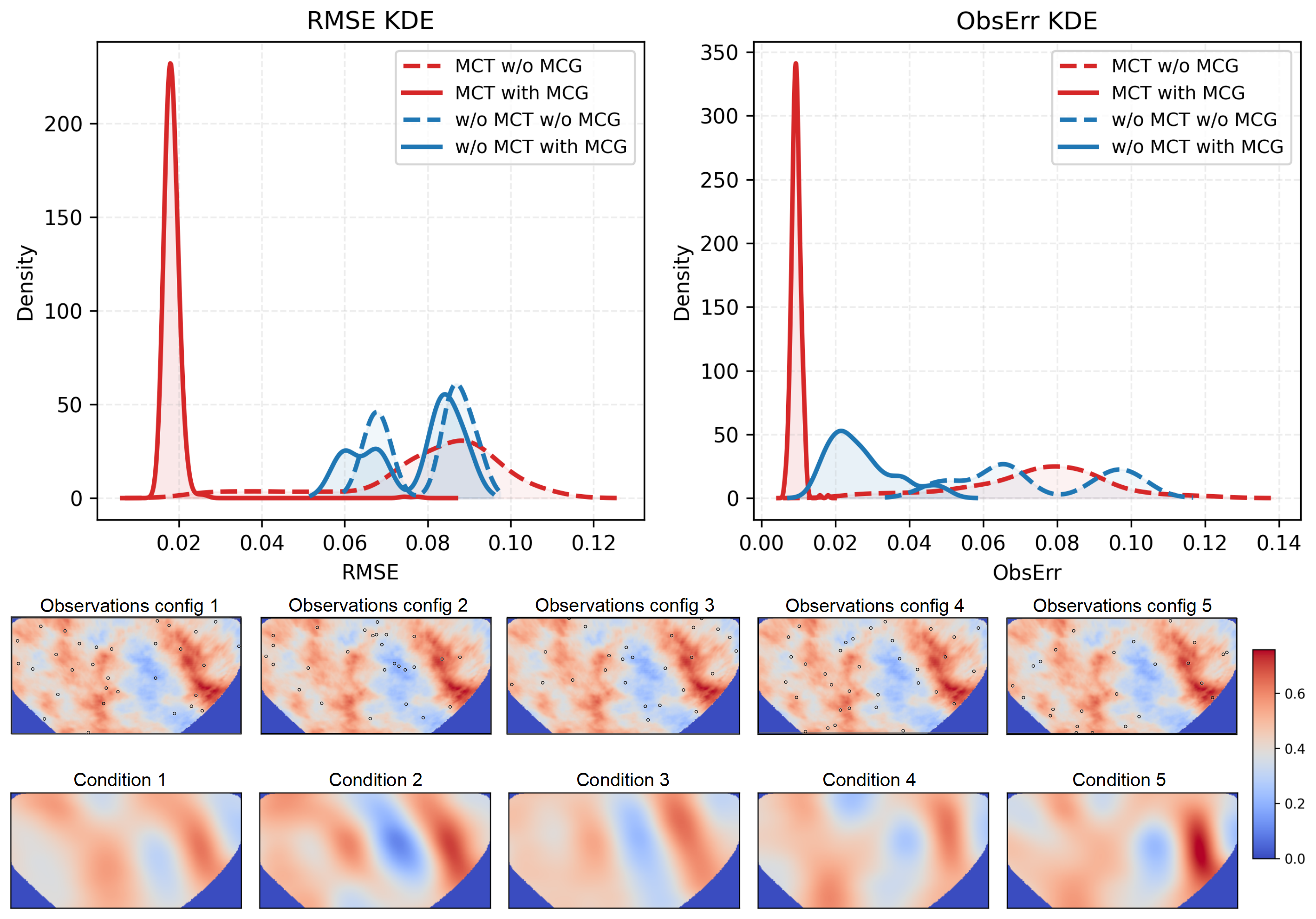}
\caption{Ablation study on the effects of Mask-Cascade Training (MCT) and Manifold-Constrained Gradient (MCG) under varying observation configurations at extreme sparsity ($0.1\%$).}
\label{fig:sea_surface_mct_comparison}
\end{figure}

\subsection{Reconstructing Navier--Stokes vorticity fields with simulation data}

To further evaluate Cas-Sensing on multiscale flow fields, we consider a two-dimensional incompressible Navier--Stokes vorticity dataset adopted from the benchmark introduced in \cite{li2020fourier}. The flow is defined on the periodic domain $\Omega=\mathbb{T}^2$ and follows the vorticity--streamfunction formulation of the incompressible Navier--Stokes equations:
\begin{equation}
\partial_t \omega
=
\hat{\boldsymbol{z}}\cdot
\left(
\nabla \times (\boldsymbol{u}\times \omega \hat{\boldsymbol{z}})
\right)
+
\nu \Delta \omega
+
\varphi,
\end{equation}
where $\omega$ denotes the scalar vorticity field, $\boldsymbol{u}$ is the incompressible velocity field, $\nu=10^{-4}$ is the viscosity coefficient, and $\varphi$ is an external forcing term given by
\begin{equation}
\varphi(\boldsymbol{x})
=
\frac{1}{10}\sin(2\pi x_1 + 2\pi x_2)
+
\frac{1}{10}\cos(2\pi x_1 + 2\pi x_2).
\end{equation}

The dataset consists of $10000$ trajectories, each initialized from a Gaussian random field and evolved using a pseudospectral solver on a $64\times64$ periodic grid. Each trajectory contains $30$ time steps, and we extract the final snapshot from each trajectory to construct a dataset of $10000$ vorticity fields for sparse reconstruction.

Before evaluating the full Cas-Sensing framework on the Navier--Stokes dataset, we first examine whether the functional autoencoder can provide a reliable coarse-scale structural condition for vorticity fields. The $10000$ vorticity snapshots are first normalized using min--max normalization, with the first $9000$ frames used for training and the remaining $1000$ frames reserved for testing. The functional autoencoder is trained with a latent dimension of 128 using a $2\%$ input point ratio. This setting follows the observation from the previous case that moderately sparse training inputs provide a better trade-off between sparse-to-coarse inference and robustness under extreme testing sparsity.

\begin{figure}[H]
\centering 
\includegraphics[width=0.8\textwidth]{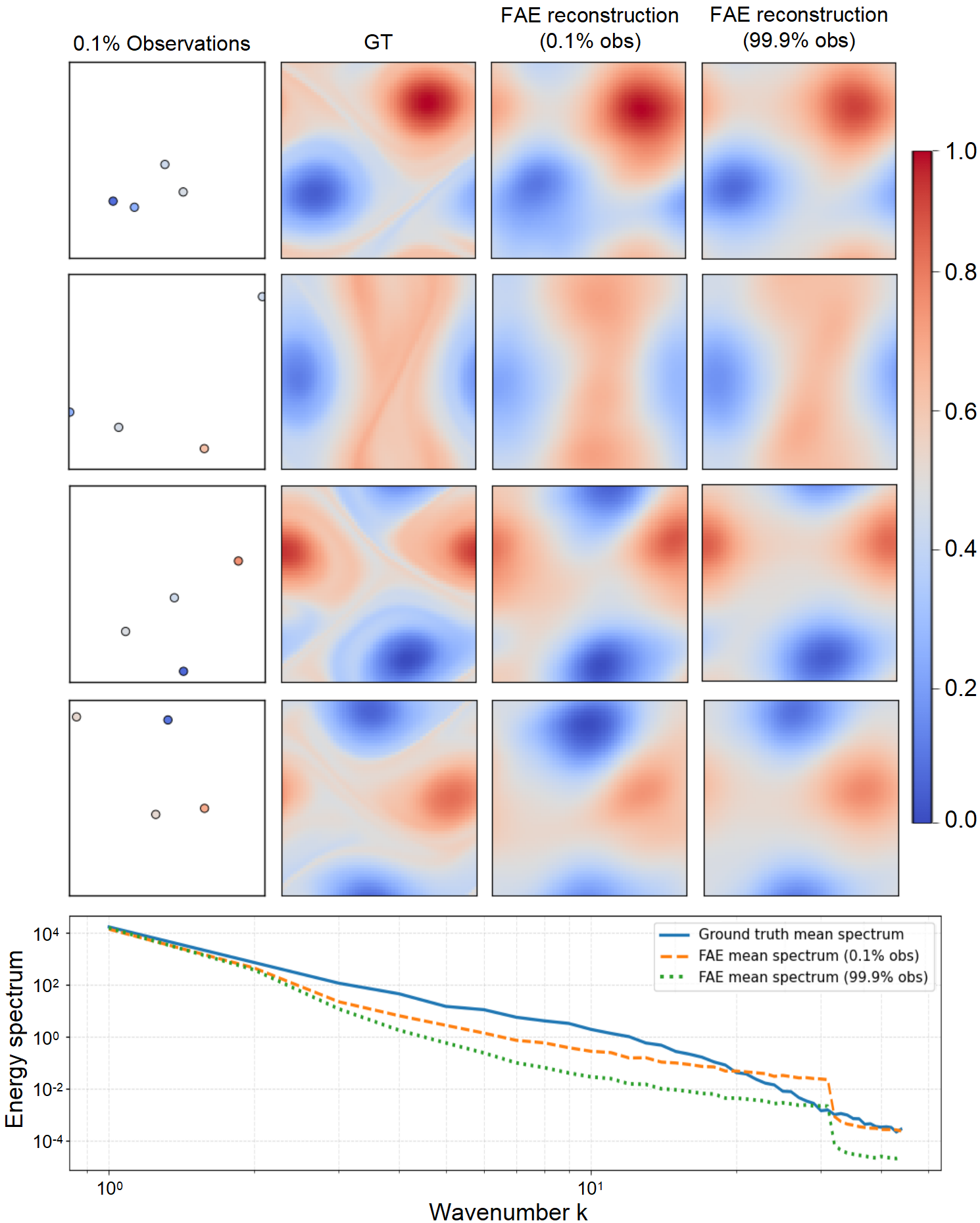}
\caption{
FAE reconstruction and energy spectrum comparison on the Navier--Stokes vorticity dataset. The FAE captures the dominant low-wavenumber structures and preserves the main spectral trend even under extremely sparse ($0.1\%$) observations.
}
\label{fig:turbulence_fae_spectrum}
\end{figure}

To assess whether the learned coarse representation preserves the physically relevant large-scale content of the flow fields, we evaluate the trained functional autoencoder under an extreme testing condition with only $0.1\%$ observations. Fig.~\ref{fig:turbulence_fae_spectrum} compares the ground-truth vorticity fields, sparse observations, FAE reconstructions, and the corresponding energy spectra. Even with only a few observed points, the FAE reconstructions recover the dominant low-wavenumber structures of the ground-truth fields, including the main large-scale vortical regions. This behavior is further confirmed by the averaged energy spectrum over representative test samples: the spectrum of the FAE reconstruction under $0.1\%$ observations closely follows the ground-truth spectrum in the low-wavenumber range, indicating that the coarse-scale energetic components are accurately captured. As expected, the high-wavenumber energy is attenuated because the FAE is designed to provide a smooth structural anchor rather than to reconstruct all fine-scale fluctuations. The dense-observation case ($99.9\%$) shows a similar low-wavenumber agreement, further confirming that the functional autoencoder mainly learns a stable low-frequency representation rather than overfitting to a particular sparse mask. These results demonstrate that, for multiscale vorticity fields with coherent vortical structures, the FAE can extract a physically meaningful coarse-scale condition that preserves the dominant spectral content needed for the subsequent conditional diffusion model.

\begin{figure}[H]
\centering 
\includegraphics[width=1\textwidth]{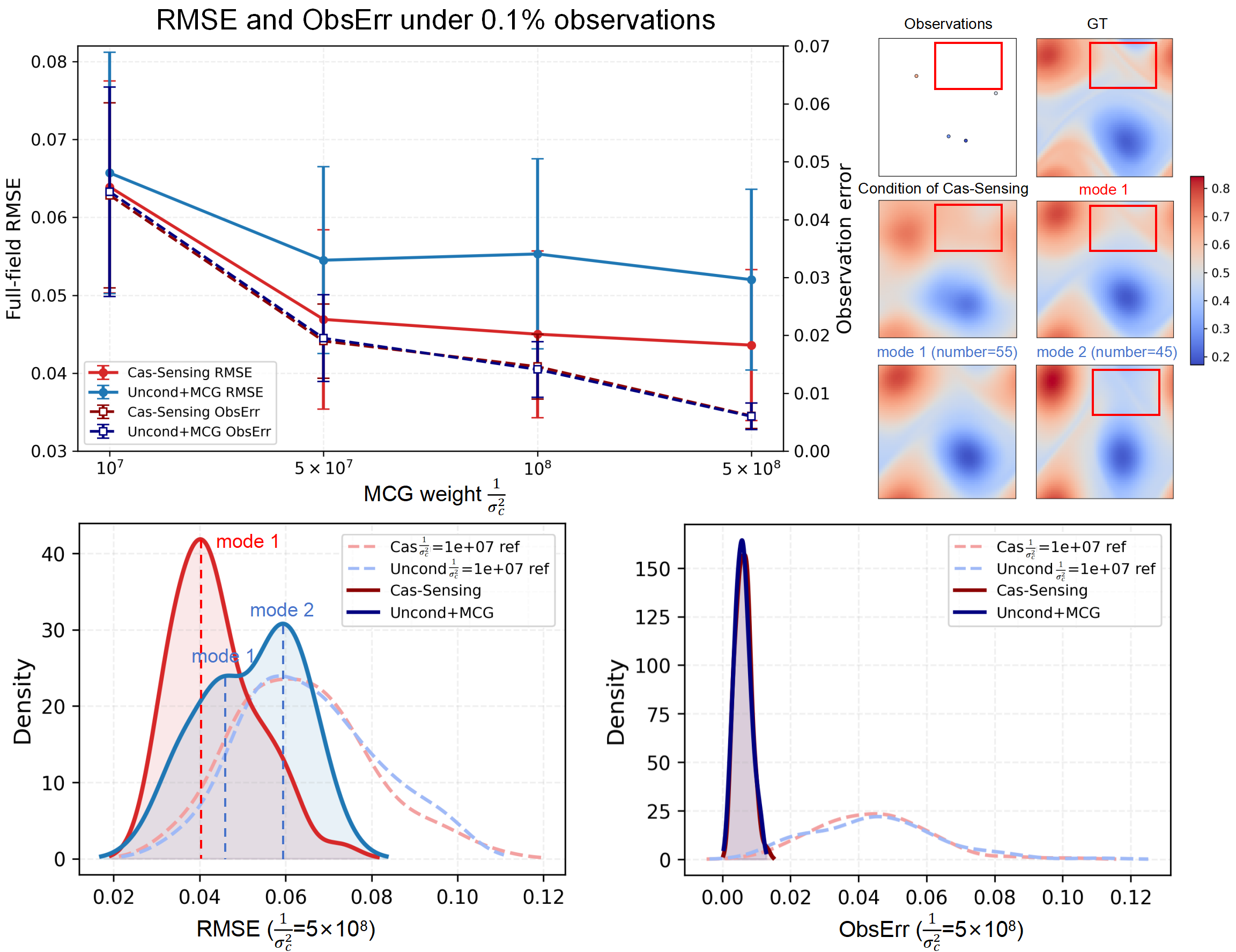}
\caption{
Comparison between Cas-Sensing and Uncond+MCG on the Navier--Stokes vorticity dataset under extremely sparse noise-free observations with varying likelihood weights.
}
\label{fig:turbulence_RMSE_ObsErr_uncond_comparison}
\end{figure}

After training the functional autoencoder, its parameters are frozen and used to generate coarse-scale structural conditions for the conditional diffusion model. The CDM is trained with Mask-Cascade Training (MCT), where $0.2\%$ of spatial points are randomly sampled and passed through the pretrained FAE to produce coarse structural estimates for residual diffusion learning. To evaluate generalization across sensor configurations, we test the model using a lower observation ratio of $0.1\%$. Fig.~\ref{fig:turbulence_RMSE_ObsErr_uncond_comparison} compares Cas-Sensing and Uncond+MCG on the Navier--Stokes vorticity dataset under this extremely sparse noise-free setting. Similar to the previous sea surface wave height field case, both methods require sufficiently large likelihood weights to effectively enforce observation consistency, and the reconstruction errors generally decrease as the MCG weight increases. Moreover, while the observation errors (ObsErr) of the two methods become nearly identical at large likelihood weights, their full-field reconstruction performance differs substantially. In particular, the RMSE gap between Cas-Sensing and Uncond+MCG gradually increases as the likelihood weight becomes larger, indicating that accurate sparse reconstruction of vorticity fields is not determined solely by fitting the observed points.

The difference becomes more evident when examining the posterior distributions at $\frac{1}{\sigma_c^2}=5\times10^8$. Although both methods produce highly concentrated ObsErr distributions, their RMSE distributions exhibit fundamentally different structures. Cas-Sensing produces a single dominant mode with concentrated RMSE statistics, indicating that the reconstructed fields remain structurally consistent across generated samples. In contrast, Uncond+MCG exhibits two clearly separated modes with comparable proportions, suggesting that multiple substantially different vorticity fields remain consistent with the sparse observations and the learned diffusion prior. The corresponding mode-mean reconstructions further reveal the origin of this behavior. In regions without observations, highlighted by the red boxes, the second mode of Uncond+MCG fails to recover the correct large-scale vortical structure and instead converges to an alternative low-frequency solution that still satisfies the sparse observations. By contrast, the coarse structural condition inferred by the functional autoencoder already predicts a physically plausible low-wavenumber structure in these unobserved regions. As a result, Cas-Sensing avoids the emergence of incorrect competing modes even under extremely sparse sensing conditions. These results further support the central motivation of Cas-Sensing: by first reconstructing a deterministic coarse-scale structural anchor and then modeling only the residual detail distribution, the solution space can be effectively constrained before posterior sampling, thereby substantially improving the stability of sparse flow-field reconstruction.

\begin{figure}[H]
\centering 
\includegraphics[width=1\textwidth]{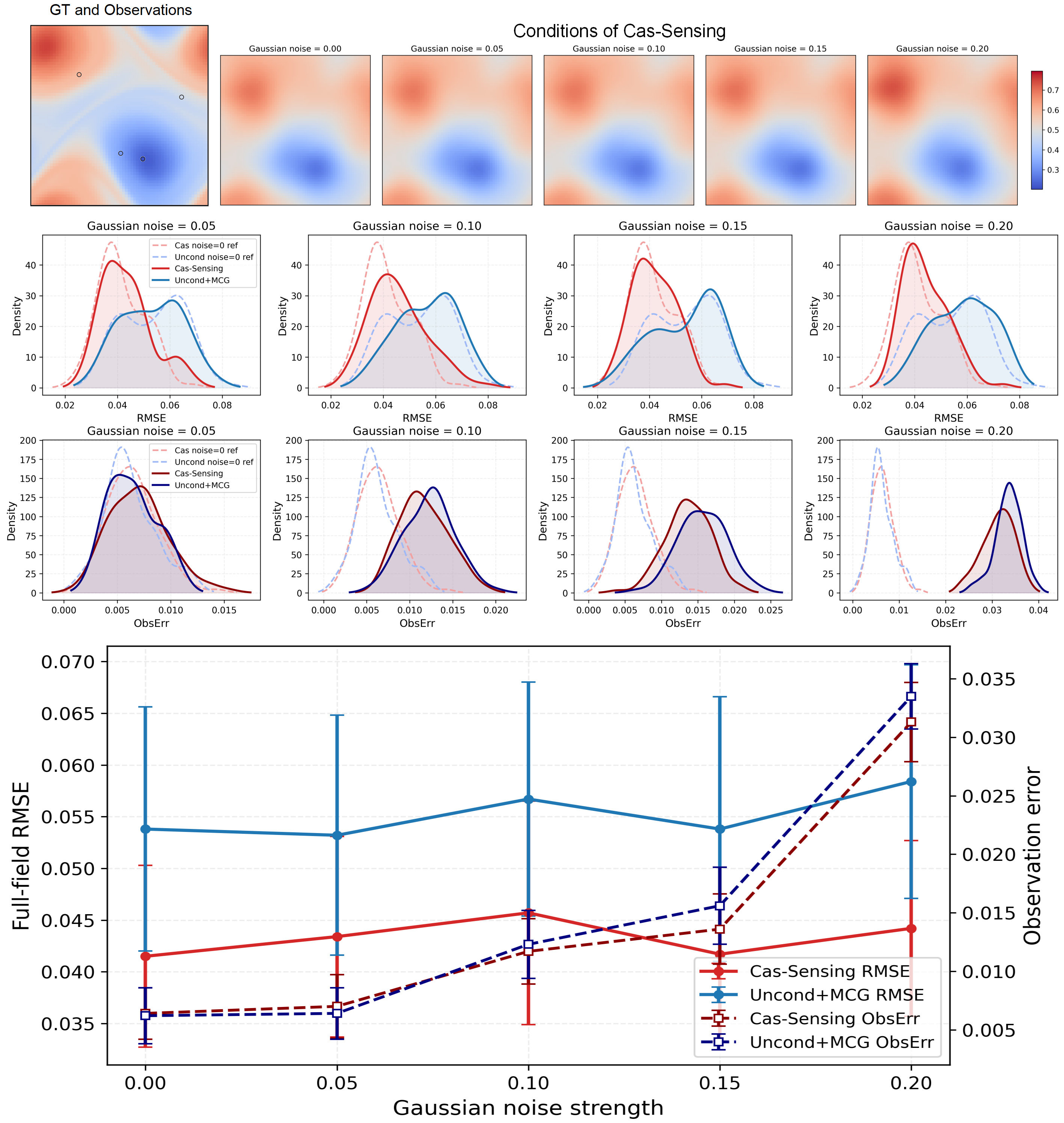}
\caption{
Comparison results of Cas-Sensing and Uncond+MCG on the Navier--Stokes vorticity dataset under varying Gaussian observation noise levels at $0.1\%$ sparsity.
}
\label{fig:turbulence_noise_comparison}
\end{figure}

We next evaluate robustness to Gaussian observation noise under extreme sparsity ($0.1\%$) with a fixed likelihood weight, as shown in Fig.~\ref{fig:turbulence_noise_comparison}. As the noise strength increases, both methods exhibit larger ObsErr, reflecting the unavoidable effect of noisy measurements on observation guidance. However, their full-field reconstruction behaviors differ markedly. Cas-Sensing maintains consistently lower RMSE across all noise levels, and its RMSE KDE remains relatively concentrated around the low-error region. In contrast, Uncond+MCG becomes increasingly unstable as the observation noise increases, with its RMSE distribution gradually shifting toward larger errors. This difference is especially pronounced at $20\%$ Gaussian noise. At this noise level, the RMSE KDE of Uncond+MCG almost completely departs from the noise-free reference distribution and collapses into a dominant high-error mode. This indicates that the inferred posterior is no longer balanced around the correct reconstruction basin; instead, the sampling process is driven toward an alternative observation-consistent but structurally incorrect solution.

Although the ObsErr distributions of the two methods remain relatively comparable, their RMSE distributions are substantially different, confirming again that satisfying sparse observations is insufficient for accurate full-field reconstruction under extreme sparsity. The robustness of Cas-Sensing can be attributed to the stability of the coarse structural condition. Across different noise levels, the FAE-generated conditions remain visually similar and preserve the dominant low-wavenumber structures of the ground-truth field. Therefore, the main structural degrees of freedom are constrained before diffusion sampling, and MCG only needs to refine the reconstruction locally. By contrast, Uncond+MCG relies entirely on noisy sparse observations to guide samples from an unconditional prior. When the observations are both sparse and noisy, this guidance becomes fragile and may steer the sampling process toward an incorrect posterior basin. These results further demonstrate that the advantage of Cas-Sensing does not come from stronger observation fitting, but from stabilizing posterior inference through a deterministic coarse-scale structural anchor. Although this MAP-like FAE approximation suppresses part of the uncertainty associated with less probable coarse configurations, it prevents noise-induced drift toward incorrect large-scale solutions, which is particularly important for practical sparse sensing under extreme sparsity and noisy measurements.

\subsection{Reconstructing cylinder flow velocity fields with simulation data}
% 先用FAE隐空间中的特征说明为什么FAE可以学到几何特征，然后展示FAE效果，最后再展示CDM效果
Having demonstrated in the previous two experiments that Cas-Sensing provides stable posterior reconstruction under extremely sparse observations, we further examine its applicability to physical systems with varying internal geometric boundaries. Such settings commonly arise in engineering problems, such as flows around bridge piers, heat-exchanger tubes, turbine blades, building layouts, internal channels, and immersed obstacles, where the flow field is strongly constrained by geometry and the boundary configurations encountered during deployment may differ from those seen during training. Therefore, this experiment is designed to assess whether Cas-Sensing can generalize to unseen boundary configurations and still reconstruct physically plausible full fields from extremely sparse observations.

We consider a synthetic two-dimensional cylinder-flow dataset to evaluate whether Cas-Sensing can be extended to physical fields with internal geometric boundaries. In contrast to the regular-domain cases considered above, the cylinder introduces geometry-dependent flow separation, vortex shedding, and wake structures. The key question is whether the functional autoencoder can infer a meaningful coarse structural condition for cylinder configurations absent from training, so that the conditional diffusion model can subsequently refine the unresolved flow details. The dataset describes incompressible flow around a circular cylinder in a two-dimensional channel and is governed by the incompressible Navier--Stokes equations:
\begin{align}
        \nabla\cdot\mathbf{u}&=0,\\
        \frac{\partial\mathbf{u}}{\partial t}+(\mathbf{u}\cdot\nabla)\mathbf{u}&=\mu\nabla^2\mathbf{u}-\frac1\rho\nabla p.
\end{align}

\begin{figure}[H]
\centering 
\includegraphics[width=0.7\textwidth]{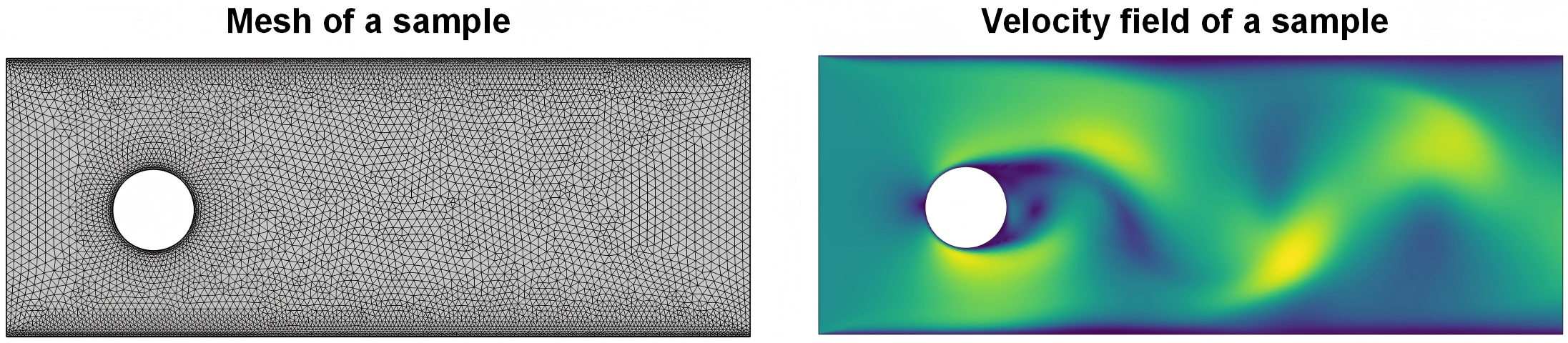}
\caption{An example of 2D cylinder flow velocity fields. The dataset contains 96 different boundary configurations of circular cylinders, each associated with 100 temporal snapshots.}
\label{fig:cylinder flow example}
\end{figure}

The flow fields are generated from transient incompressible laminar-flow simulations in a $1.6 \times 0.6~\mathrm{m}$ rectangular channel. Geometric variability is introduced by Latin hypercube sampling of the cylinder center and radius, with $x_c\in[0.25,0.65]~\mathrm{m}$, $y_c\in[0.15,0.45]~\mathrm{m}$, and $R\in[0.05,0.10]~\mathrm{m}$. The fluid density and dynamic viscosity are $\rho=1.5~\mathrm{kg/m^3}$ and $\mu=5\times10^{-4}~\mathrm{Pa\cdot s}$, respectively. A uniform inlet velocity of $U=2.0~\mathrm{m/s}$ is prescribed, with zero static pressure at the outlet and no-slip conditions on the walls and cylinder surface. Taking $D=2R$ as the characteristic length, the Reynolds number $Re=\rho U D/\mu$ ranges from $600$ to $1200$, leading to unsteady wake flows with periodic vortex shedding. The transient simulations use an implicit BDF scheme with adaptive time stepping, initialized with a time step of $0.001~\mathrm{s}$. After post-processing, the dataset used in this study contains 96 boundary configurations, each with 100 temporal snapshots interpolated onto a $256 \times 96$ uniform grid. A representative sample is shown in Fig.~\ref{fig:cylinder flow example}.

For the purpose of model training and evaluation, the dataset is partitioned based on boundary configurations rather than temporal snapshots, ensuring that the test set contains \textit{unseen} geometrical configurations. Specifically, 76 boundary configurations, corresponding to a total of 7600 flow field snapshots, are employed as the training set, while the remaining 20 configurations, consisting of 2000 snapshots, are reserved exclusively for testing.

We first employ a functional autoencoder to learn compact representations of the circular cylinder flow fields. To encourage robustness, the model is trained in a masked manner, where only $2\%$ of the grid points in each snapshot are provided to the encoder and the task is to reconstruct the full field from this partial information. 

\begin{figure}[H]
\centering 
\includegraphics[width=1\textwidth]{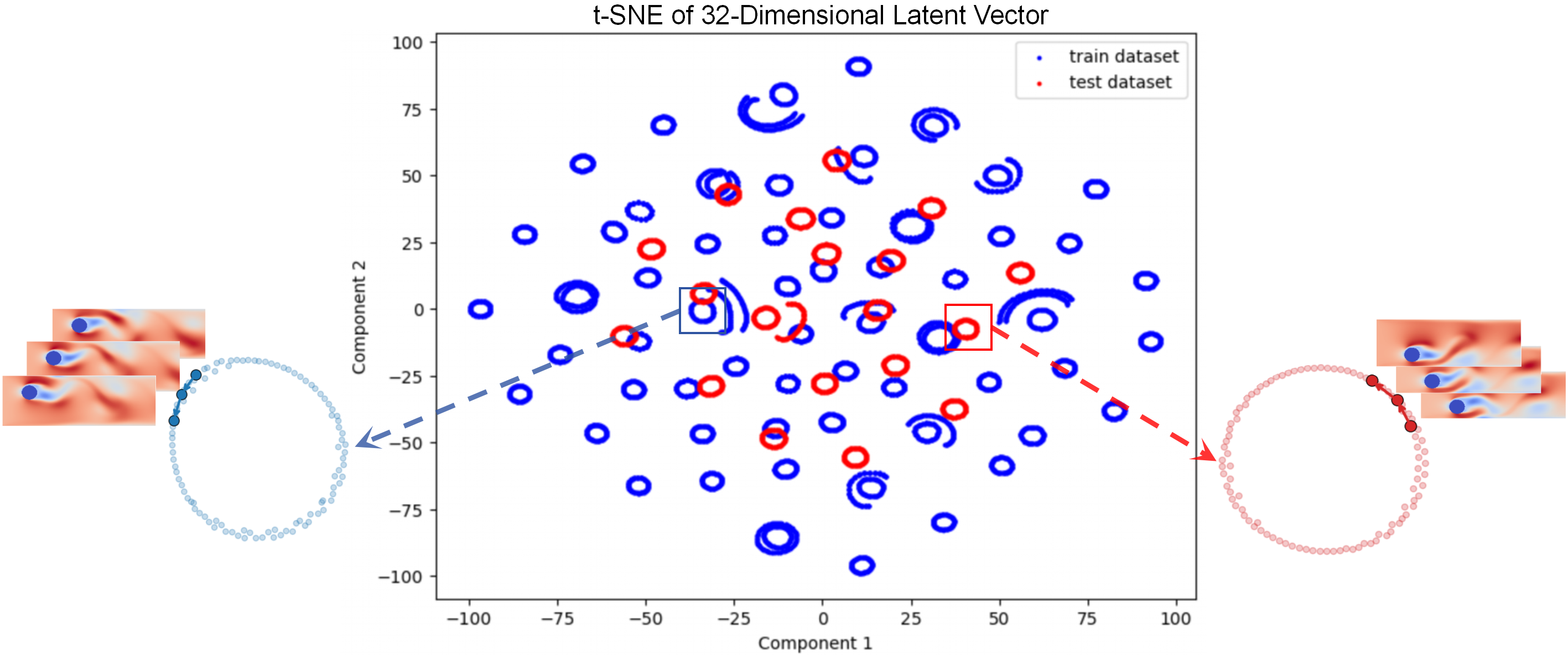}
\caption{t-SNE visualization of latent representations learned by the functional autoencoder for cylinder flow fields.}
\label{fig:cylinder flow latent space}
\end{figure}

To investigate the latent representation learned by the functional autoencoder, we first visualize the extracted latent vectors of the cylinder flow dataset. Specifically, the 32-dimensional latent vectors are projected into a two-dimensional space using the t-distributed Stochastic Neighbor Embedding (t-SNE) technique \cite{maaten2008visualizing}. Each scattered point in Fig. \ref{fig:cylinder flow latent space} corresponds to the latent variable of a single flow snapshot. It can be observed that the training data associated with different internal boundary configurations are well separated in the latent space, while the latent vectors of snapshots from the same configuration form circular patterns. This circular structure reflects the inherent periodicity of the velocity fields in cylinder wake flows. Moreover, the latent vectors corresponding to the test dataset exhibit the same characteristic structures, indicating that the functional autoencoder has effectively captured both the temporal and geometric features of the given flow fields.

\begin{figure}[H]
\centering 
\includegraphics[width=1.\textwidth]{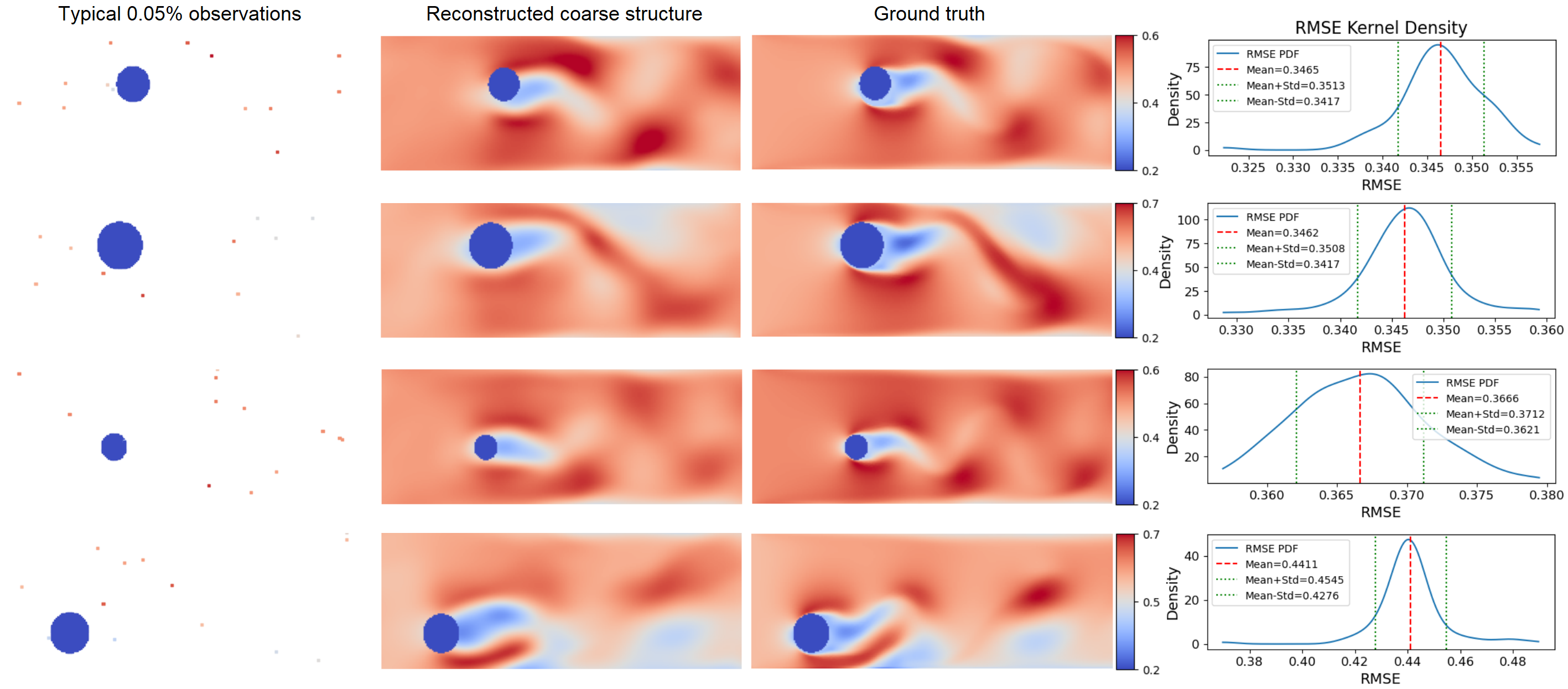}
\caption{The reconstructed coarse-scale structures and kernel density estimates for RMSE on the reference sample across 100 randomly chosen meshes.}
\label{fig:cylinder flow FAE ratio comparision}
\end{figure}

To further assess the capability of the functional autoencoder under extreme sparsity and unseen geometric boundary conditions, we select a representative flow field from the test set and apply 100 randomly generated masks with sampling ratios of only $0.05\%$. Since all test samples correspond to cylinder geometries that are never observed during training, this experiment directly evaluates whether the functional autoencoder can generalize to unseen boundary-induced flow structures from extremely sparse measurements. The masked inputs are mapped through the trained functional autoencoder to reconstruct the complete field on the $256\times96$ grid, and the resulting RMSE distributions are shown in Fig.~\ref{fig:cylinder flow FAE ratio comparision}. Despite the extremely limited observations, the reconstructed coarse-scale fields still recover the dominant wake topology, large-scale flow organization, and boundary-induced structures associated with the unseen cylinder geometries. Although fine-scale discrepancies remain near the cylinder boundary and high-gradient wake regions, the inferred coarse structures remain physically reasonable and structurally consistent with the ground truth. These results indicate that the neural-operator-based functional autoencoder learns a transferable geometry-aware sparse-to-structure mapping rather than memorizing geometry-specific flow realizations. More importantly, such reconstructed coarse-scale fields provide reliable structural anchors for the downstream conditional diffusion model, substantially reducing the difficulty of subsequent fine-scale generative refinement under unseen geometric conditions.

\begin{figure}[H]
\centering 
\includegraphics[width=1.\textwidth]{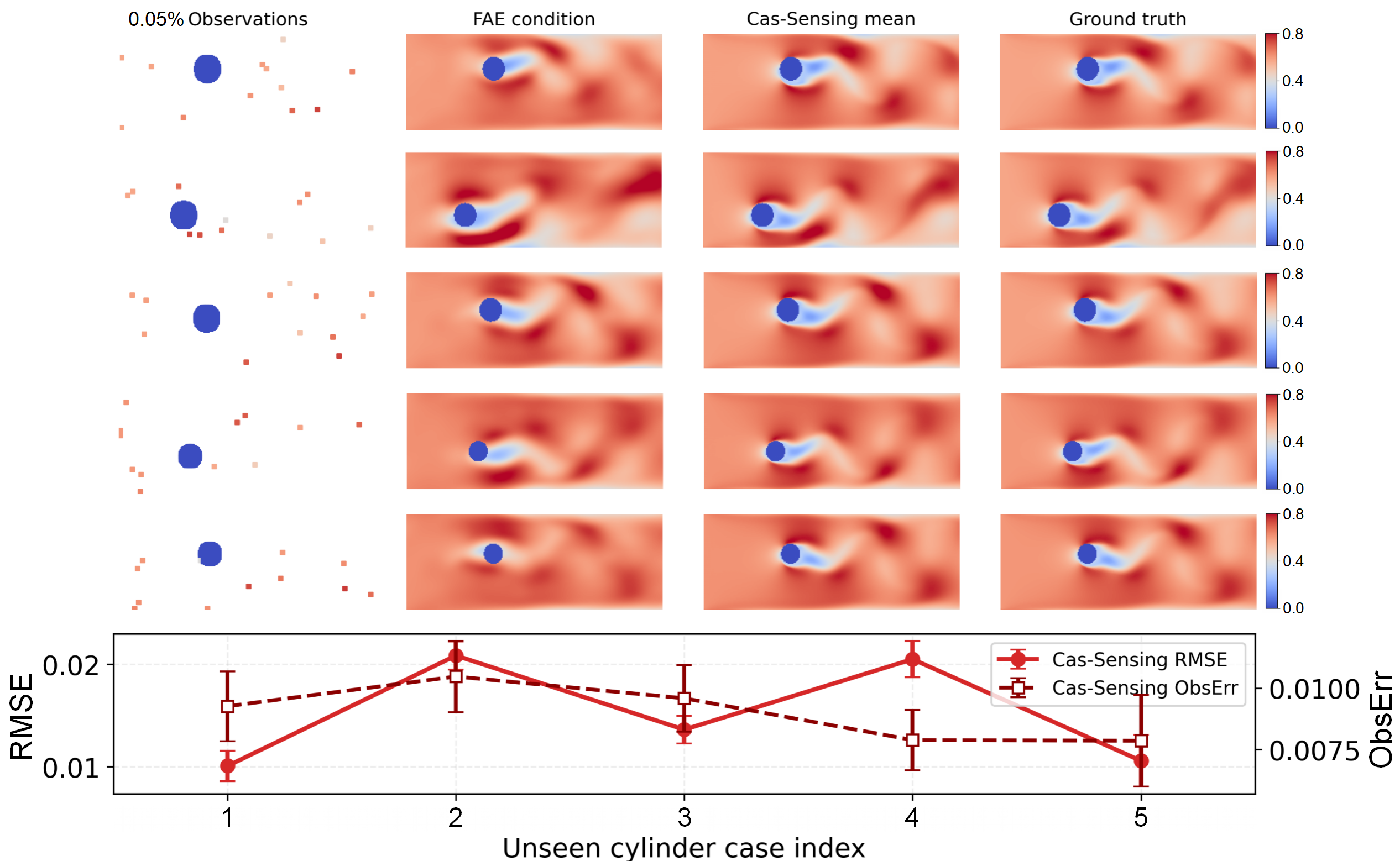}
\caption{The full-field reconstruction results of cylinder flow fields with $0.05\%$ observations and unseen geometries.}
\label{fig:cylinder_flow_CDM_0.05}
\end{figure}

After training the functional autoencoder, its parameters are frozen and used to provide coarse-scale structural conditions for the conditional diffusion model (CDM). To evaluate whether Cas-Sensing can be extended to physical fields with unseen internal geometric boundaries, we train the CDM using the same training split containing only the first 76 cylinder geometries, and then test it on cylinder configurations that are absent from the training set. During mask-cascade training, $0.1\%$ sparse measurements are supplied to the functional autoencoder, which reconstructs geometry-aware coarse flow structures as conditions for residual diffusion learning. During inference, these coarse structural conditions are further combined with the available measurements through manifold-constrained guidance to recover the complete flow fields, with the MCG likelihood weight set to $1/\sigma_c^2 = 10^7$. For each unseen test case, we generate 100 independent reconstructions and report the mean reconstruction and the corresponding error statistics. Representative results under only $0.05\%$ observations are shown in Fig.~\ref{fig:cylinder_flow_CDM_0.05}. Across different unseen cylinder configurations, the FAE conditions already capture the main wake organization and boundary-induced flow structures, providing meaningful structural anchors for the CDM. The final Cas-Sensing mean reconstructions closely match the corresponding ground-truth fields, including the obstacle location, near-cylinder flow separation, and downstream wake patterns. The quantitative results further show consistently low full-field RMSE and observation error across the tested unseen cases. These results indicate that Cas-Sensing can generalize beyond the geometric configurations observed during training and can be applied to sparse reconstruction problems involving varying internal boundaries.

\subsection{Reconstructing global sea surface temperature fields with reanalysis data}

\begin{figure}[H]
\centering 
\includegraphics[width=0.8\textwidth]{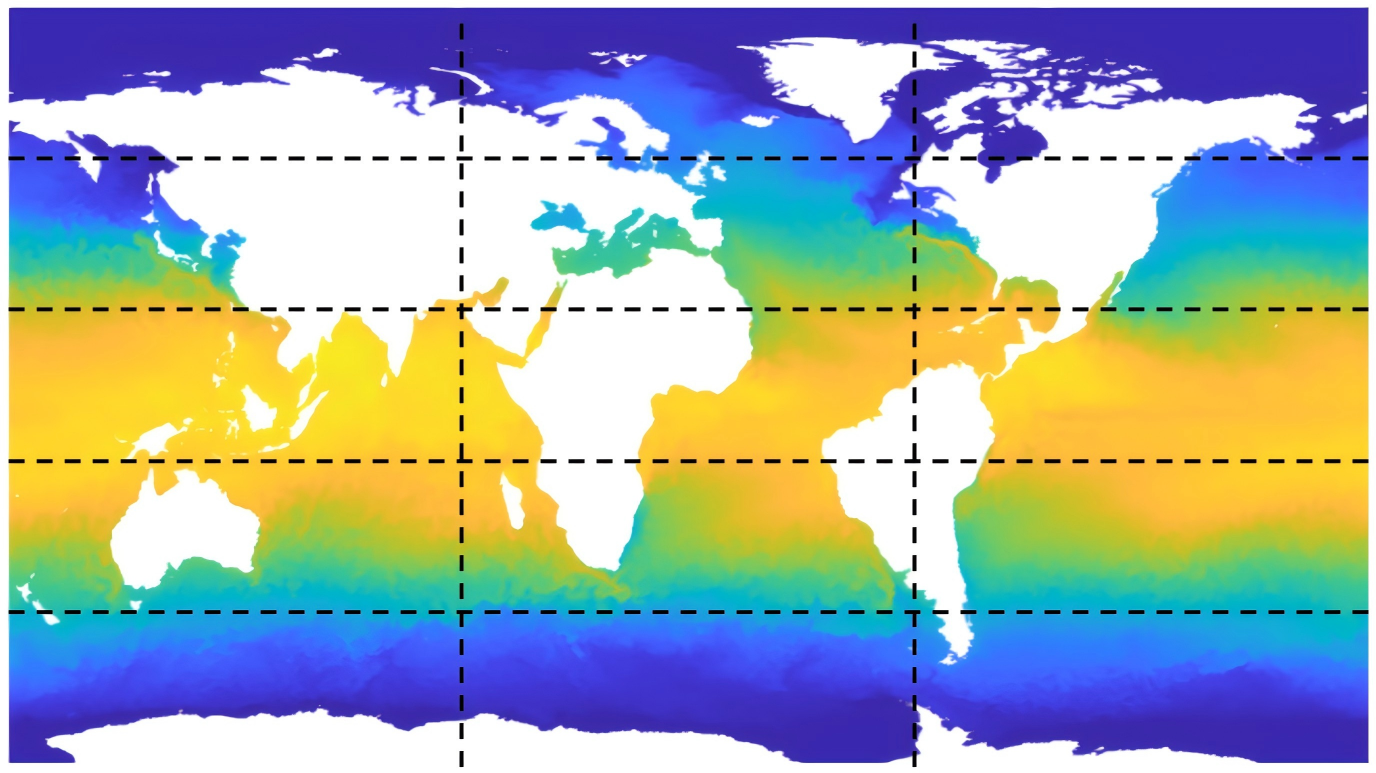}
\caption{Partition of the global sea surface temperature field dataset.}
\label{fig:global sea surface temp example}
\end{figure}

We further apply Cas-Sensing to a larger-scale real-world physical system using the daily global Sea Surface Temperature (SST) reanalysis dataset provided by the Copernicus Marine Service \cite{SST_dataset}. This experiment is designed to evaluate whether the proposed framework can be extended from controlled benchmark fields to global geophysical reconstruction problems, where the target field spans a much larger spatial domain and contains complex land--sea boundaries. We selected data covering three years, from June 1, 2022, to May 30, 2025. Due to the high spatial resolution of the original dataset ($4320 \times 2040$), the fields were first downsampled to $1440 \times 640$. The downsampled global fields were then evenly partitioned into 15 subregions, each with a resolution of $480 \times 128$, as illustrated in Fig.~\ref{fig:global sea surface temp example}. These subregions together form the SST reconstruction dataset used in this study. All subregions are reconstructed using a shared FAE and a shared CDM, rather than training separate models for different regions, to assess the ability of Cas-Sensing to handle diverse geophysical subdomains with common model parameters. For model training and evaluation, the first $80\%$ of the temporal samples were used for training, and the remaining $20\%$ were reserved for testing.

\begin{figure}[H]
\centering 
\subfigure[Ensemble-mean full-field reconstruction under $0.1\%$ observations, obtained from 100 independent generations for each subregion.]{
\label{fig:global sea temp CDM 0.1}
\includegraphics[width=1.0\textwidth]{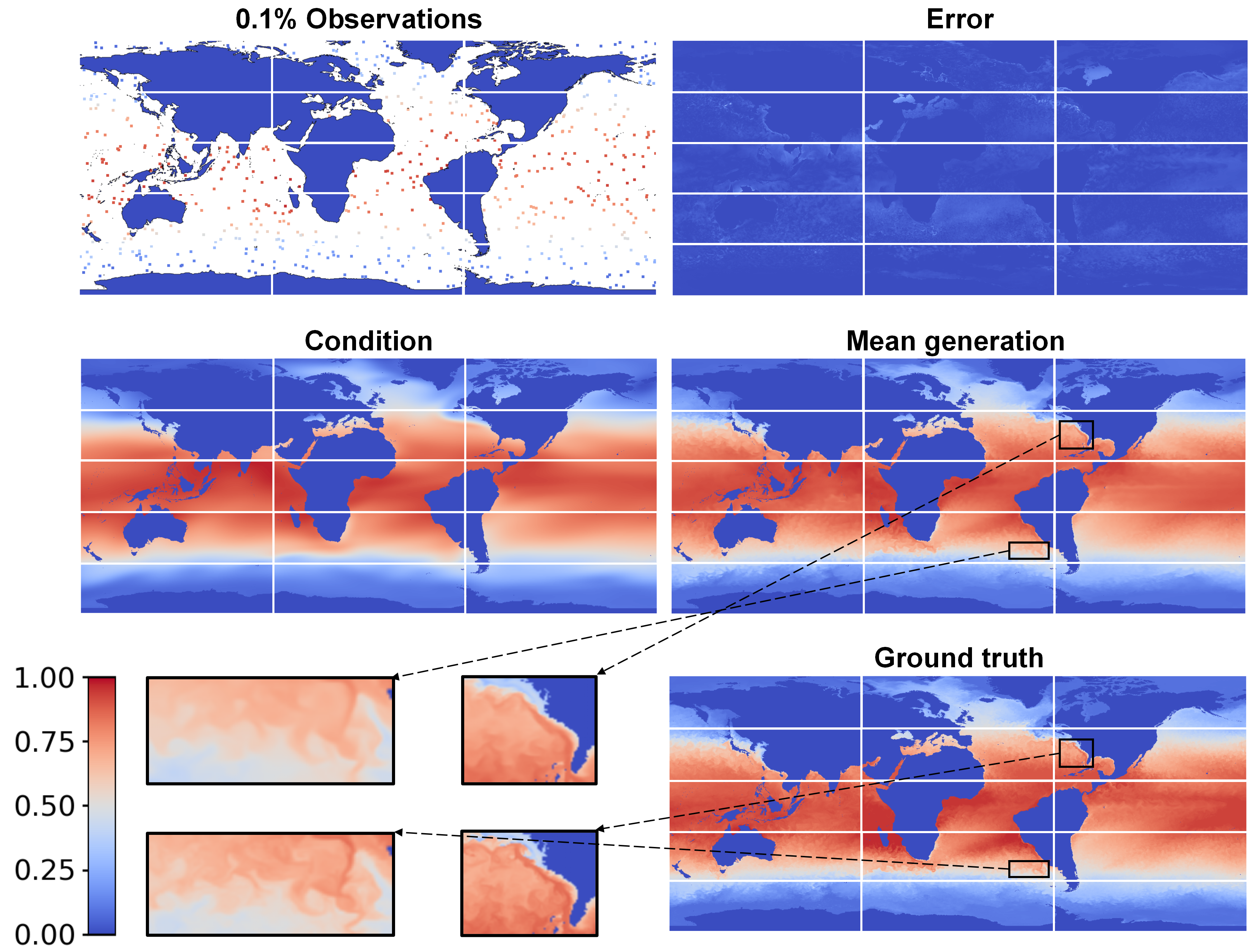}}
\subfigure[Ensemble-mean reconstructions from 100 generations under different input point ratios, together with the corresponding RMSE statistics. The sparse mask is fixed for each input ratio.]{
\label{fig:global sea temp CDM ratio comparision}
\includegraphics[width=1.0\textwidth]{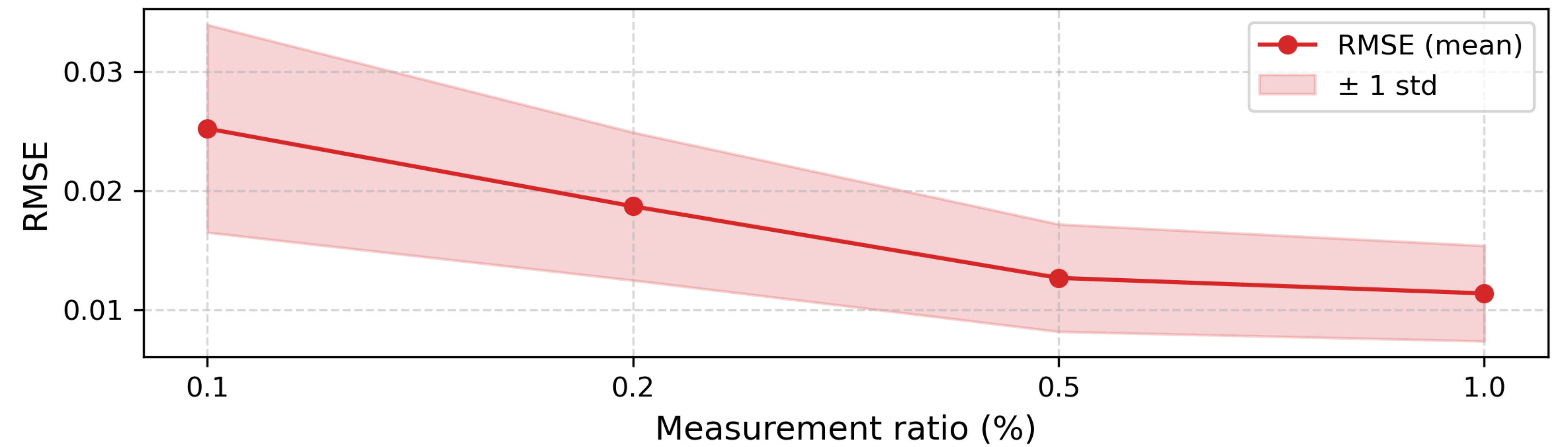}}
\caption{The full-field reconstruction results of global sea temperature fields.}
\label{fig:Cas-Sensing global sea temp results}
\end{figure}

We begin by training a functional autoencoder on the dataset to learn a compact representation of the global sea surface temperature field. During the masked training, $1\%$ of the data points in each frame are randomly selected as inputs to the encoder, while the remaining $99\%$ are used as targets for the decoder. As illustrated in Fig. \ref{fig:global sea temp CDM 0.1} (\textit{Condition}), the trained functional autoencoder is capable of accurately reconstructing the coarse-scale background patterns from only partial observations, capturing the essential spatial variability of the ocean surface temperature fields. Once the autoencoder is fully trained, it is frozen and integrated into the Cascaded Sensing framework to provide background field reconstructions that serve as conditioning information for the conditional diffusion model during mask-cascade training. Specifically, a sparse subset of points ($0.2\%$) is randomly sampled at each step during training, from which the functional autoencoder reconstructs a smooth background field. The conditional diffusion model then learns to refine this background and recover the full field. After 800 epochs of training, the conditional diffusion model is well-trained to reconstruct refined-scale details based on conditions and extremely sparse measurements.

To evaluate the model's generalization ability under more challenging observation conditions, we randomly sample only $0.1\%$ of the data points from the 15 predefined global subregions, a sampling ratio which is half of the sampling ratio used during training. During inference, manifold-constrained guidance is applied with the likelihood weight set to $1/\sigma_c^2 = 10^7$. For each subregion, we generate 100 independent reconstructions and report their ensemble mean as the final reconstruction. The averaged global reconstruction result is shown in Fig.~\ref{fig:global sea temp CDM 0.1}. Despite the extremely limited observations, Cas-Sensing accurately recovers the large-scale global SST distribution and preserves coherent regional temperature patterns across different subregions. As shown in the zoomed-in views, the reconstructed fields also retain refined local structures, indicating that the learned residual prior can provide meaningful fine-scale corrections beyond the coarse FAE estimate even under severe observation sparsity.

To systematically assess the robustness of the proposed approach, we further conduct experiments under a range of observation ratios ($1\%, 0.5\%, 0.2\%, 0.1\%$). For each ratio, sparse points are randomly selected and fixed, and the model generates 100 samples to characterize the reconstruction variability. The resulting mean and standard deviation of RMSE are summarized in Fig. \ref{fig:global sea temp CDM ratio comparision}. As expected, both the mean and standard deviation of RMSE increase as the observation ratio decreases. Nevertheless, the errors remain consistently low across all tested ratios, underscoring the robustness, stability, and generalization capability of the cascaded sensing framework in extremely sparse sensing scenarios.

\section{Discussion and conclusion}\label{sec:discussion and conclusion}

In this work, we presented Cascaded Sensing (Cas-Sensing), a hierarchical probabilistic framework for reconstructing physical fields from extremely sparse measurements. The central message of this study is that sparse scientific sensing should not be treated merely as deterministic interpolation or one-step conditional generation. Under extreme sparsity, the posterior over full fields is severely underconstrained and sensitive to observation patterns and noise. The key challenge is therefore not only to represent uncertainty, but to organize it in a stable and structural manner.

Cas-Sensing addresses this challenge by restructuring posterior inference into a coarse-to-detail cascade. Instead of asking a single model to learn or sample the full posterior $p(\boldsymbol{u}\mid\boldsymbol{y})$ directly, the framework first uses a neural-operator-based functional autoencoder to infer a deterministic coarse structural estimate from sparse measurements. This estimate acts as a structural anchor that fixes the dominant global degrees of freedom of the reconstruction. The remaining task is then converted into a residual inference problem, where a conditional diffusion model learns a detail prior conditioned on the inferred coarse structure. In this way, the generative stage is confined to a better-conditioned neighborhood around a plausible global configuration, rather than being forced to resolve the entire ambiguity of the original inverse problem.

The experimental results support this interpretation. In the sea-surface wave reconstruction experiments, the comparison with unconditional diffusion with MCG shows that matching sparse observations alone is not sufficient for accurate full-field reconstruction. Even when observation errors become comparable, the unconditional model can still fluctuate among structurally different full-field solutions, while Cas-Sensing produces more concentrated and accurate posterior samples. The trajectory analysis further shows that, without a coarse structural anchor, MCG may act as a global mode-selection force and split the reverse diffusion trajectories, whereas in Cas-Sensing it mainly serves as a local refinement mechanism. Robustness tests under noisy observations and varying sensor configurations further demonstrate that the structural anchor stabilizes posterior inference against perturbations in the observation process.

The ablation study clarifies the complementary roles of the two key components in the second stage. Mask-Cascade Training (MCT) exposes the conditional diffusion model to a family of imperfect but plausible coarse conditions induced by random sparse samplings, thereby broadening and stabilizing the support of the learned detail prior. Manifold-Constrained Gradient (MCG), in contrast, enforces consistency with the actual sparse observations at inference time. The results show that MCG alone cannot recover accurate solutions if the learned distribution does not contain the correct reconstruction modes. Conversely, when MCT has already expanded the conditional support, MCG can effectively select and refine observation-consistent samples. This confirms that the advantage of Cas-Sensing comes from the combination of structured prior learning and observation-guided refinement, rather than from stronger measurement fitting alone.

The additional experiments also clarify the scope of the proposed decomposition. For turbulent Navier--Stokes vorticity fields, the functional autoencoder preserves the dominant low-wavenumber energetic structures even under $0.1\%$ observations, while attenuating high-wavenumber fluctuations as expected for a coarse structural estimator. This indicates that the coarse stage is not restricted to globally smooth fields, but can extract physically meaningful large-scale content from multiscale turbulent data. For cylinder-flow reconstruction with unseen geometric configurations, the FAE provides transferable geometry-aware coarse conditions, enabling Cas-Sensing to reconstruct wake structures under boundary configurations absent from training. Finally, the global SST experiment shows that the same shared FAE and CDM can be applied to multiple large-scale geophysical subregions with complex land--sea boundaries, suggesting that the framework can be extended beyond controlled benchmark domains to larger real-world physical systems.

These findings also reveal several important practical implications. First, the deterministic coarse stage should not be interpreted as a claim that the true coarse posterior is exactly degenerate. Rather, it is a modeling choice that compresses the relatively lower-uncertainty structural component into a tractable point estimate, while the remaining variability is reintroduced through MCT and refined through MCG. Second, the success of Cas-Sensing depends on whether the coarse structural component is sufficiently learnable from sparse observations. When the dominant structure itself becomes highly ambiguous, or when the spatial correlation length is too short for sparse sensors to constrain even the coarse scales, the deterministic anchor may become unreliable. In such regimes, a fully probabilistic coarse-stage model or ensemble-based coarse inference may be necessary. Third, the effectiveness of the residual diffusion stage depends on the diversity of coarse conditions encountered during MCT. If the training masks produce too narrow a family of conditions, the learned residual prior may not generalize well to more severe sensing conditions.

Several limitations remain. The current implementation focuses mainly on two-dimensional spatial fields and does not explicitly model temporal dynamics. Extending the framework to spatiotemporal reconstruction would require incorporating temporal priors or recurrent/transformer-based dynamics into the cascade. In addition, the present diffusion sampler is based on a standard DDPM formulation, which can be computationally expensive when many posterior samples are required. Accelerated diffusion samplers, flow-matching models, or consistency-model-based refinements may improve sampling efficiency. For the global SST experiment, the current partitioned reconstruction does not explicitly impose inter-region boundary consistency after recombining the 15 subregions. A more scalable solution would be to train the generative stage on a global or latent global representation, so that boundary continuity is handled directly by the model rather than by post hoc stitching.

Overall, Cas-Sensing provides a structured probabilistic paradigm for scientific sensing under extreme data sparsity. Its main contribution is not simply the combination of an autoencoder and a diffusion model, but the redistribution of uncertainty across stages: deterministic structural inference resolves the dominant global ambiguity, mask-cascade training exposes the generative model to condition variability, and manifold-constrained guidance enforces measurement consistency as a refinement step. This staged organization transforms an ill-conditioned full-field posterior into a sequence of better-conditioned inference problems, leading to stable and physically consistent reconstructions across sparse sensor layouts, noise perturbations, turbulent multiscale fields, unseen geometric boundaries, and large-scale geophysical domains. The results suggest that explicitly structuring uncertainty is a promising direction for robust scientific sensing and for future data-driven inverse modeling in complex physical systems.
\section*{Acknowledgements}
The authors wish to express their gratitude for the financial support received from Guangdong Provincial Fund - Special Innovation Project (2024KTSCX038); Research Grants Council of Hong Kong through the Research Impact Fund (R5006-23); the Guangdong Provincial Key Lab of Integrated Communication, Sensing and Computation for Ubiquitous Internet of Things (No. 2023B1212010007). The authors also thank Mr. Xiao DI for generously providing the circular cylinder flow dataset used in this study.
\section*{Appendix. Implementation Details}\label{sec:appendix}

Additional implementation details of the functional autoencoder (FAE) and conditional diffusion model (CDM) used in different experiments are summarized in the following tables. The CDM uses a U-Net backbone with 128 base channels, channel multipliers [1, 2, 3, 4], two residual blocks per scale, and a learning rate of \(1\times10^{-5}\) for all experiments. The detailed architectures of the neural-operator-based functional autoencoder are already described in Section. \ref{sec:methods}. Unless otherwise specified, all models were trained using the Adam optimizer on a single NVIDIA RTX PRO 6000 GPU. Code and processed datasets will be released upon publication. 

\begin{table}[htbp]
\centering
\caption{Training details of the functional autoencoder (FAE).}
\label{tab:fae_details}
\small
\begin{tabular}{lcccccc}
\hline
Dataset & Latent dim & Batch size & LR & Epochs & Masked input ratio & Training time (h)\\
\hline
Sea-surface wave height & 128 & 4 & 1e-3 & 40 & $2\%$ & 0.40\\
Navier--Stokes vorticity & 128 & 4 & 1e-3 & 40 & $2\%$ & 0.42 \\
Cylinder flow & 32 & 4 & 1e-3 & 50 & $1\%$ & 0.36\\
Global sea surface temperature & 64 & 4 & 1e-3 & 80 & $1\%$ & 0.97\\
\hline
\end{tabular}
\end{table}

\begin{table}[htbp]
\centering
\caption{Training details of the conditional diffusion model (CDM) used in different experiments.}
\label{tab:cdm_details}
\small
\begin{tabular}{lcccc}
\hline
Dataset & Epochs & Batch size & MCT input ratio & Training time (h) \\
\hline
Sea-surface wave height & 1000 & 32 & $0.2\%$ & 20.45 \\
Navier--Stokes vorticity & 1000 & 32 & $0.2\%$ & 8.80 \\
Cylinder flow & 1000 & 50 & $0.1\%$ & 25.12 \\
Global sea surface temperature& 800 & 30 & $0.2\%$ & 37.25 \\
\hline
\end{tabular}
\end{table}
% \bibliography{sn-bibliography}% common bib file

\end{document}